\documentclass[]{bytedance_seed}

\usepackage[toc,page,header]{appendix}

\usepackage{xcolor}
\usepackage{amssymb}
\usepackage{amsmath}
\usepackage{mathtools}
\usepackage{booktabs}
\usepackage{multirow}
\usepackage{colortbl}
\usepackage{tikz}
\usepackage{dsfont}
\usetikzlibrary{arrows.meta, positioning, decorations.pathreplacing, calc, fit, backgrounds, shadows}
\definecolor{encoderBlue}{RGB}{66, 133, 244}
\definecolor{encoderLight}{RGB}{232, 240, 254}
\definecolor{decoderPurple}{RGB}{142, 68, 173}
\definecolor{decoderLight}{RGB}{245, 238, 248}
\definecolor{thinkOrange}{RGB}{230, 126, 34}
\definecolor{thinkLight}{RGB}{253, 245, 230}
\definecolor{conceptGreen}{RGB}{39, 174, 96}
\definecolor{conceptLight}{RGB}{232, 248, 239}
\definecolor{inputTeal}{RGB}{0, 150, 136}
\definecolor{inputLight}{RGB}{224, 242, 241}
\definecolor{outputGold}{RGB}{241, 196, 15}
\definecolor{outputLight}{RGB}{254, 249, 231}
\definecolor{rose}{RGB}{231, 76, 60}
\definecolor{violet}{RGB}{155, 89, 182}
\definecolor{gray200}{RGB}{229, 231, 235}
\definecolor{gray400}{RGB}{156, 163, 175}
\definecolor{gray600}{RGB}{75, 85, 99}
\definecolor{tabgray}{gray}{0.95} 
\newcommand{\ModelName}{DLCM}

\title{ Dynamic Large Concept Models: Latent Reasoning in an Adaptive Semantic Space }

\affiliation[1]{ByteDance Seed}
\affiliation[2]{University of Manchester}
\affiliation[3]{Mila - Quebec AI Institute}
\affiliation[4]{Tsinghua University }
\affiliation[5]{M-A-P}

\abstract{
Large Language Models (LLMs) apply uniform computation to all tokens, despite language exhibiting highly non-uniform information density. This token-uniform regime wastes capacity on locally predictable spans while under-allocating computation to semantically critical transitions. We propose \textbf{Dynamic Large Concept Models (DLCM)}, a hierarchical language modeling framework that learns semantic boundaries from latent representations and shifts computation from tokens to a compressed concept space where reasoning is more efficient. DLCM discovers variable-length concepts end-to-end without relying on predefined linguistic units. Hierarchical compression fundamentally changes scaling behavior. We introduce the first \textbf{compression-aware scaling law}, which disentangles token-level capacity, concept-level reasoning capacity, and compression ratio, enabling principled compute allocation under fixed FLOPs. To stably train this heterogeneous architecture, we further develop a \textbf{decoupled $\mu$P parametrization} that supports zero-shot hyperparameter transfer across widths and compression regimes. At a practical setting ($R=4$, corresponding to an average of four tokens per concept), DLCM reallocates roughly one-third of inference compute into a higher-capacity reasoning backbone, achieving a \textbf{+2.69\% average improvement} across 12 zero-shot benchmarks under matched inference FLOPs.

}

\date{\today}
\correspondence{\email{xingwei.qu@bytedance.com},
\email{zhangge.eli@bytedance.com}}

\begin{document}
\maketitle


\section{Introduction}

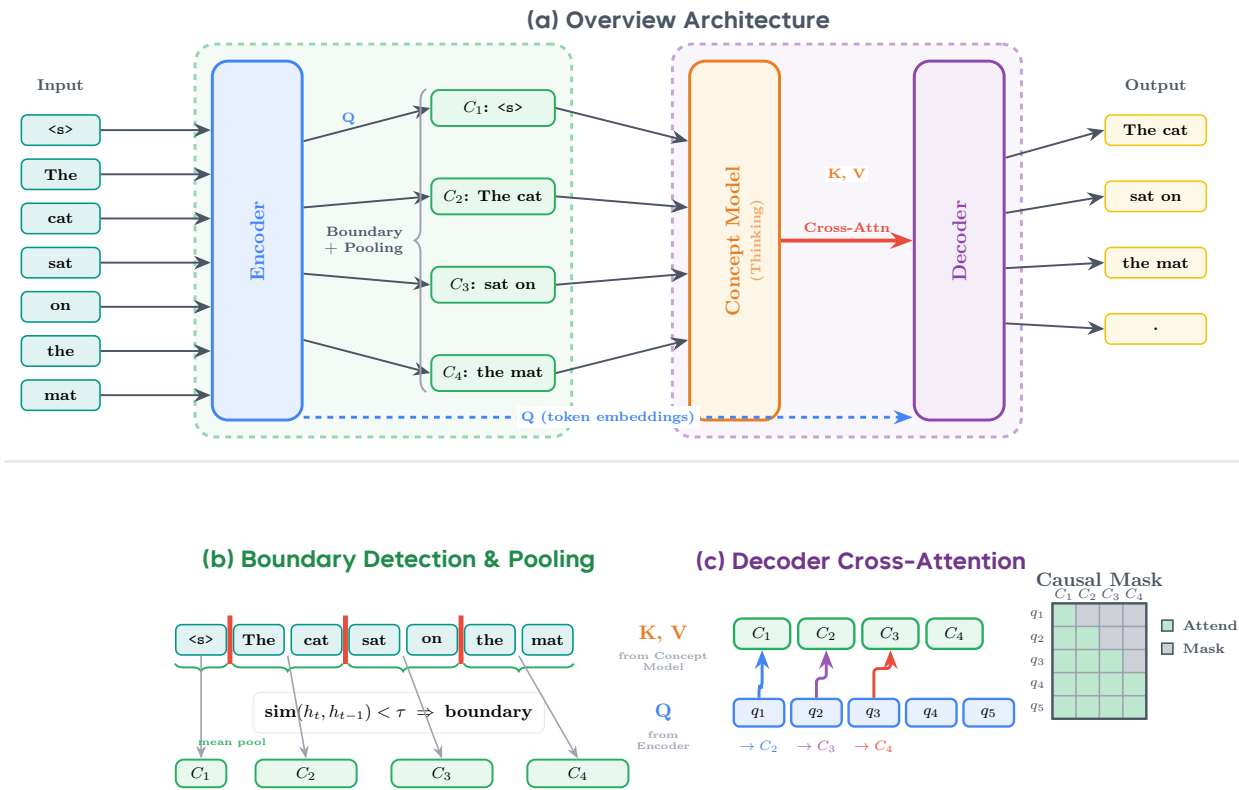
\begin{figure}[htbp]
\centering
\resizebox{\textwidth}{0.46\textheight}{%
\begin{tikzpicture}[
    >=Stealth,
    every node/.style={font=\small\bfseries},
    module/.style={
        rounded corners=8pt,
        line width=1.5pt,
        minimum width=1.6cm,
        minimum height=5cm,
        align=center,
        inner sep=8pt
    },
    concept/.style={
        draw=conceptGreen,
        fill=conceptLight,
        rounded corners=4pt,
        font=\footnotesize\bfseries,
        minimum width=2.2cm,
        minimum height=0.65cm,
        align=center,
        line width=1pt
    },
    inputtoken/.style={
        draw=inputTeal,
        fill=inputLight,
        rounded corners=3pt,
        font=\footnotesize\bfseries,
        minimum width=1.4cm,
        minimum height=0.55cm,
        align=center,
        line width=0.8pt
    },
    outputtoken/.style={
        draw=outputGold,
        fill=outputLight,
        rounded corners=3pt,
        font=\footnotesize\bfseries,
        minimum width=1.8cm,
        minimum height=0.55cm,
        align=center,
        line width=0.8pt
    },
    arrow/.style={
        -{Stealth[length=3mm, width=2.5mm]},
        line width=1pt,
        draw=gray600
    },
    thickarrow/.style={
        -{Stealth[length=4mm, width=3mm]},
        line width=2pt
    },
    title/.style={
        font=\large\bfseries,
        text=gray600
    },
    sublabel/.style={
        font=\scriptsize\bfseries,
        text=gray400
    }
]


\node[title] at (7.5, 7) {\textsf{(a) Overview Architecture}};

\def\midY{3}

\node[font=\footnotesize\bfseries, text=gray600] at (-3.5, \midY+2.8) {Input};

\node[inputtoken] (i0) at (-3.5, \midY+2.0) {\texttt{<s>}};
\node[inputtoken] (i1) at (-3.5, \midY+1.2) {The};
\node[inputtoken] (i2) at (-3.5, \midY+0.4) {cat};
\node[inputtoken] (i3) at (-3.5, \midY-0.4) {sat};
\node[inputtoken] (i4) at (-3.5, \midY-1.2) {on};
\node[inputtoken] (i5) at (-3.5, \midY-2.0) {the};
\node[inputtoken] (i6) at (-3.5, \midY-2.8) {mat};

\node[module, draw=encoderBlue, fill=encoderLight, minimum height=6.5cm] (enc) at (0, \midY) {};
\node[rotate=90, font=\bfseries, text=encoderBlue] at (enc.center) {Encoder};

\foreach \i in {0,1,2,3,4,5,6} {
    \draw[arrow] (i\i.east) -- ([yshift={2.0cm-0.8*\i cm}]enc.west);
}

\node[concept] (c1) at (4.2, \midY+2.4) {$C_1$: \texttt{<s>}};
\node[concept] (c2) at (4.2, \midY+0.8) {$C_2$: The cat};
\node[concept] (c3) at (4.2, \midY-0.8) {$C_3$: sat on};
\node[concept] (c4) at (4.2, \midY-2.4) {$C_4$: the mat};

\node[module, draw=thinkOrange, fill=thinkLight, minimum height=6.5cm] (cmodel) at (8.5, \midY) {};
\node[rotate=90, font=\bfseries, text=thinkOrange, align=center] at (cmodel.center) {Concept Model};
\node[rotate=90, font=\scriptsize\bfseries, text=thinkOrange!70] at ([xshift=4mm]cmodel.center) {(Thinking)};

\node[module, draw=decoderPurple, fill=decoderLight, minimum height=6.5cm] (dec) at (12.5, \midY) {};
\node[rotate=90, font=\bfseries, text=decoderPurple] at (dec.center) {Decoder};

\node[font=\footnotesize\bfseries, text=gray600] at (16, \midY+2.8) {Output};

\node[outputtoken] (o1) at (16, \midY+2.0) {The cat};
\node[outputtoken] (o2) at (16, \midY+0.8) {sat on};
\node[outputtoken] (o3) at (16, \midY-0.4) {the mat};
\node[outputtoken] (o4) at (16, \midY-1.6) {.};

\node[font=\scriptsize\bfseries, text=encoderBlue] at ([xshift=0.8cm, yshift=2.2cm]enc.east) {Q};
\draw[arrow] ([yshift=1.8cm]enc.east) -- (c1.west);
\draw[arrow] ([yshift=0.6cm]enc.east) -- (c2.west);
\draw[arrow] ([yshift=-0.6cm]enc.east) -- (c3.west);
\draw[arrow] ([yshift=-1.8cm]enc.east) -- (c4.west);

\draw[arrow] (c1.east) -- ([yshift=1.8cm]cmodel.west);
\draw[arrow] (c2.east) -- ([yshift=0.6cm]cmodel.west);
\draw[arrow] (c3.east) -- ([yshift=-0.6cm]cmodel.west);
\draw[arrow] (c4.east) -- ([yshift=-1.8cm]cmodel.west);

\draw[thickarrow, draw=rose] (cmodel.east) -- node[above, font=\scriptsize\bfseries, text=rose] {Cross-Attn} (dec.west);
\node[font=\scriptsize\bfseries, text=thinkOrange, fill=white, inner sep=1pt] at (10.5, \midY+1.2) {K, V};

\draw[->, encoderBlue, line width=1.5pt, dashed] 
    ([yshift=-3.2cm]enc.east) -- ([yshift=-3.2cm]dec.west);
\node[font=\scriptsize\bfseries, text=encoderBlue, fill=white, inner sep=2pt] at (6.25, \midY-3.2) {Q (token embeddings)};

\draw[arrow] ([yshift=1.5cm]dec.east) -- (o1.west);
\draw[arrow] ([yshift=0.5cm]dec.east) -- (o2.west);
\draw[arrow] ([yshift=-0.5cm]dec.east) -- (o3.west);
\draw[arrow] ([yshift=-1.5cm]dec.east) -- (o4.west);

\draw[decorate, decoration={brace, amplitude=6pt, raise=2pt}, gray400, line width=1pt] 
    (c4.south west) -- (c1.north west) 
    node[midway, left=10pt, font=\scriptsize\bfseries, text=gray600, align=right] {Boundary\\+ Pooling};

\begin{scope}[on background layer]
    \node[draw=conceptGreen!50, dashed, line width=1.5pt, rounded corners=10pt, 
          fit=(enc) (c1) (c4), inner sep=8pt, fill=conceptGreen!5] (boxB) {};
\end{scope}

\begin{scope}[on background layer]
    \node[draw=decoderPurple!50, dashed, line width=1.5pt, rounded corners=10pt, 
          fit=(cmodel) (dec), inner sep=8pt, fill=decoderPurple!5] (boxC) {};
\end{scope}

\draw[gray200, line width=1.5pt] (-4.5, -1) -- (17.5, -1);


\begin{scope}[shift={(-1, -5.0)}, scale=1.1]
    \node[title, text=conceptGreen!80!black] at (3.2, 2.0) {\textsf{(b) Boundary Detection \& Pooling}};
    
    \node[inputtoken, minimum width=0.9cm] (t0) at (0, 0.7) {\texttt{<s>}};
    \node[inputtoken, minimum width=0.9cm, right=0.1cm of t0] (t1) {The};
    \node[inputtoken, minimum width=0.9cm, right=0.1cm of t1] (t2) {cat};
    \node[inputtoken, minimum width=0.9cm, right=0.1cm of t2] (t3) {sat};
    \node[inputtoken, minimum width=0.9cm, right=0.1cm of t3] (t4) {on};
    \node[inputtoken, minimum width=0.9cm, right=0.1cm of t4] (t5) {the};
    \node[inputtoken, minimum width=0.9cm, right=0.1cm of t5] (t6) {mat};
    
    \draw[rose, line width=2.5pt] ($(t0.east)!0.5!(t1.west)$) ++(0, 0.4) -- ++(0, -0.8);
    \draw[rose, line width=2.5pt] ($(t2.east)!0.5!(t3.west)$) ++(0, 0.4) -- ++(0, -0.8);
    \draw[rose, line width=2.5pt] ($(t4.east)!0.5!(t5.west)$) ++(0, 0.4) -- ++(0, -0.8);
    
    \node[draw=gray200, fill=white, rounded corners=4pt, inner sep=6pt, font=\small\bfseries] at (3.2, -0.5) 
        {$\text{sim}(h_t, h_{t-1}) < \tau \;\Rightarrow\; \text{boundary}$};
    
    \draw[decorate, decoration={brace, amplitude=4pt, mirror, raise=4pt}, conceptGreen, line width=1pt] 
        (t0.south west) -- (t0.south east);
    \draw[decorate, decoration={brace, amplitude=4pt, mirror, raise=4pt}, conceptGreen, line width=1pt] 
        (t1.south west) -- (t2.south east);
    \draw[decorate, decoration={brace, amplitude=4pt, mirror, raise=4pt}, conceptGreen, line width=1pt] 
        (t3.south west) -- (t4.south east);
    \draw[decorate, decoration={brace, amplitude=4pt, mirror, raise=4pt}, conceptGreen, line width=1pt] 
        (t5.south west) -- (t6.south east);
    
    \node[concept, minimum width=0.9cm, minimum height=0.5cm] (pc1) at (0, -1.5) {$C_1$};
    \node[concept, minimum width=1.8cm, minimum height=0.5cm] (pc2) at (1.7, -1.5) {$C_2$};
    \node[concept, minimum width=1.8cm, minimum height=0.5cm] (pc3) at (3.9, -1.5) {$C_3$};
    \node[concept, minimum width=1.8cm, minimum height=0.5cm] (pc4) at (6.1, -1.5) {$C_4$};
    
    \draw[->, gray400, line width=0.8pt] (t0.south) -- (pc1.north);
    \draw[->, gray400, line width=0.8pt] ($(t1.south)!0.5!(t2.south)$) -- (pc2.north);
    \draw[->, gray400, line width=0.8pt] ($(t3.south)!0.5!(t4.south)$) -- (pc3.north);
    \draw[->, gray400, line width=0.8pt] ($(t5.south)!0.5!(t6.south)$) -- (pc4.north);
    
    \node[font=\tiny\bfseries, text=conceptGreen] at (0.5, -1.0) {mean pool};
\end{scope}


\begin{scope}[shift={(8.0, -5.0)}, scale=1.1]
    \node[title, text=decoderPurple!80!black] at (2.5, 2.0) {\textsf{(c) Decoder Cross-Attention}};
    
    \node[font=\normalsize\bfseries, text=thinkOrange] at (-0.7, 0.8) {K, V};
    \node[font=\tiny\bfseries, text=gray400, align=center] at (-0.7, 0.35) {from Concept\\Model};
    
    \node[concept, minimum width=1.0cm, minimum height=0.55cm, font=\footnotesize\bfseries, line width=1.2pt] (kv1) at (0.9, 0.8) {$C_1$};
    \node[concept, minimum width=1.0cm, minimum height=0.55cm, font=\footnotesize\bfseries, line width=1.2pt, right=0.1cm of kv1] (kv2) {$C_2$};
    \node[concept, minimum width=1.0cm, minimum height=0.55cm, font=\footnotesize\bfseries, line width=1.2pt, right=0.1cm of kv2] (kv3) {$C_3$};
    \node[concept, minimum width=1.0cm, minimum height=0.55cm, font=\footnotesize\bfseries, line width=1.2pt, right=0.1cm of kv3] (kv4) {$C_4$};
    
    \node[font=\normalsize\bfseries, text=encoderBlue] at (-0.7, -0.5) {Q};
    \node[font=\tiny\bfseries, text=gray400, align=center] at (-0.7, -0.95) {from\\Encoder};
    
    \node[draw=encoderBlue, fill=encoderLight, rounded corners=3pt, minimum width=0.9cm, minimum height=0.5cm, font=\footnotesize\bfseries, line width=1.2pt] (q1) at (0.85, -0.5) {$q_1$};
    \node[draw=encoderBlue, fill=encoderLight, rounded corners=3pt, minimum width=0.9cm, minimum height=0.5cm, font=\footnotesize\bfseries, line width=1.2pt, right=0.08cm of q1] (q2) {$q_2$};
    \node[draw=encoderBlue, fill=encoderLight, rounded corners=3pt, minimum width=0.9cm, minimum height=0.5cm, font=\footnotesize\bfseries, line width=1.2pt, right=0.08cm of q2] (q3) {$q_3$};
    \node[draw=encoderBlue, fill=encoderLight, rounded corners=3pt, minimum width=0.9cm, minimum height=0.5cm, font=\footnotesize\bfseries, line width=1.2pt, right=0.08cm of q3] (q4) {$q_4$};
    \node[draw=encoderBlue, fill=encoderLight, rounded corners=3pt, minimum width=0.9cm, minimum height=0.5cm, font=\footnotesize\bfseries, line width=1.2pt, right=0.08cm of q4] (q5) {$q_5$};
    
    \draw[->, encoderBlue, line width=1.5pt, rounded corners=2pt] (q1.north) -- ++(0, 0.18) -| (kv1.south);
    \draw[->, violet, line width=1.5pt, rounded corners=2pt] (q2.north) -- ++(0, 0.32) -| (kv2.south);
    \draw[->, rose, line width=1.5pt, rounded corners=2pt] (q3.north) -- ++(0, 0.48) -| (kv3.south);
    
    \node[font=\scriptsize\bfseries, text=encoderBlue] at (q1.south) [below=3pt] {$\to C_2$};
    \node[font=\scriptsize\bfseries, text=violet] at (q2.south) [below=3pt] {$\to C_3$};
    \node[font=\scriptsize\bfseries, text=rose] at (q3.south) [below=3pt] {$\to C_4$};
    
    \begin{scope}[shift={(5.6, -0.6)}]
        \def\cellsize{0.38}
        
        \node[font=\normalsize\bfseries, text=gray600] at (2*\cellsize, 5*\cellsize + 0.4) {Causal Mask};
        
        \fill[gray400!50] (0, 0) rectangle (4*\cellsize, 5*\cellsize);
        
        \fill[conceptGreen!30] (0, 4*\cellsize) rectangle (1*\cellsize, 5*\cellsize);
        \fill[conceptGreen!30] (0, 3*\cellsize) rectangle (2*\cellsize, 4*\cellsize);
        \fill[conceptGreen!30] (0, 2*\cellsize) rectangle (3*\cellsize, 3*\cellsize);
        \fill[conceptGreen!30] (0, 1*\cellsize) rectangle (4*\cellsize, 2*\cellsize);
        \fill[conceptGreen!30] (0, 0) rectangle (4*\cellsize, 1*\cellsize);
        
        \draw[gray400, line width=0.8pt] (0, 0) grid[step=\cellsize] (4*\cellsize, 5*\cellsize);
        \draw[gray600, line width=1.2pt] (0, 0) rectangle (4*\cellsize, 5*\cellsize);
        
        \node[font=\scriptsize\bfseries, text=gray600] at (0.5*\cellsize, 5*\cellsize+0.15) {$C_1$};
        \node[font=\scriptsize\bfseries, text=gray600] at (1.5*\cellsize, 5*\cellsize+0.15) {$C_2$};
        \node[font=\scriptsize\bfseries, text=gray600] at (2.5*\cellsize, 5*\cellsize+0.15) {$C_3$};
        \node[font=\scriptsize\bfseries, text=gray600] at (3.5*\cellsize, 5*\cellsize+0.15) {$C_4$};
        
        \node[font=\scriptsize\bfseries, text=gray600] at (-0.24, 4.5*\cellsize) {$q_1$};
        \node[font=\scriptsize\bfseries, text=gray600] at (-0.24, 3.5*\cellsize) {$q_2$};
        \node[font=\scriptsize\bfseries, text=gray600] at (-0.24, 2.5*\cellsize) {$q_3$};
        \node[font=\scriptsize\bfseries, text=gray600] at (-0.24, 1.5*\cellsize) {$q_4$};
        \node[font=\scriptsize\bfseries, text=gray600] at (-0.24, 0.5*\cellsize) {$q_5$};
        
        \fill[conceptGreen!30] (4*\cellsize+0.25, 3.8*\cellsize) rectangle (4*\cellsize+0.43, 3.8*\cellsize+0.18);
        \draw[gray600, line width=0.8pt] (4*\cellsize+0.25, 3.8*\cellsize) rectangle (4*\cellsize+0.43, 3.8*\cellsize+0.18);
        \node[font=\scriptsize\bfseries, text=gray600, right] at (4*\cellsize+0.48, 3.8*\cellsize+0.09) {Attend};
        
        \fill[gray400!50] (4*\cellsize+0.25, 2.8*\cellsize) rectangle (4*\cellsize+0.43, 2.8*\cellsize+0.18);
        \draw[gray600, line width=0.8pt] (4*\cellsize+0.25, 2.8*\cellsize) rectangle (4*\cellsize+0.43, 2.8*\cellsize+0.18);
        \node[font=\scriptsize\bfseries, text=gray600, right] at (4*\cellsize+0.48, 2.8*\cellsize+0.09) {Mask};
    \end{scope}
\end{scope}

\end{tikzpicture}
}
\caption{Overview Structure of \ModelName. }
\label{fig:architecture}
\end{figure}

Large Language Models (LLMs) have achieved remarkable success across natural language understanding, reasoning, and generation tasks. Despite differences in scale and training data, nearly all state-of-the-art models share a common architectural assumption: language is processed uniformly at the token level, with identical depth and computation applied to every position in the sequence.

This assumption stands in sharp contrast to the structure of natural language. Information density is highly non-uniform: long spans of locally predictable tokens are interspersed with sparse but semantically critical transitions where new concepts are introduced and reasoning difficulty concentrates. Yet standard LLMs expend full computation on both regimes alike, resulting in substantial redundancy and systematic misallocation of model capacity.

More fundamentally, this inefficiency reflects a limitation of token-level modeling itself. Reasoning is inherently hierarchical: humans reason over abstract units such as ideas or concepts before committing to surface realizations. Token-level autoregressive models, however, lack any explicit abstraction mechanism and are forced to repeatedly infer high-level structure implicitly at every layer, solely through next-token prediction.

Prior work has explored relaxing this constraint, but with important limitations. Latent reasoning approaches perform inference in continuous hidden spaces without explicit token generation, while sentence-level concept models rely on fixed, human-defined segmentation. Neither enables models to learn where semantic computation should be concentrated.

We argue that effective reasoning requires a learned intermediate granularity: neither raw tokens nor predefined sentences, but variable-length semantic concepts discovered directly from representation space. Based on this insight, we propose \textbf{Dynamic Large Concept Models (DLCM)}, a hierarchical next-token prediction framework that dynamically segments token sequences into concepts, performs deep reasoning in a compressed concept space, and reconstructs token-level predictions through causal cross-attention.

This design separates what to reason about from how to reason. By learning semantic boundaries end-to-end and relocating computation from redundant token processing to concept-level reasoning, DLCM enables adaptive compute allocation aligned with information density.

At the other extreme, H-NET~\cite{hwang2025dynamicchunkingendtoendhierarchical} demonstrates the promise of learned boundary detection with adaptive compute allocation, but operates at the byte level and has not been validated against standard Next Token Prediction (NTP) baselines in modern LLM pipelines.

Our work bridges these gaps by introducing \textbf{latent reasoning at the concept level}. Throughout this paper, we use the term \emph{concept} to denote variable-length latent segments discovered from representation space, rather than linguistically predefined semantic units.
 The key insight is that effective reasoning requires neither token-level granularity (too fine, computationally wasteful) nor sentence-level granularity (too coarse, inflexible), but rather semantically coherent concepts whose boundaries are learned end-to-end from data. We propose \textbf{\ModelName}, a hierarchical architecture that implements this insight through a four-stage pipeline (Figure~\ref{fig:architecture}):

\begin{enumerate}
\item \textbf{Encoding:} A lightweight encoder processes raw tokens to extract fine-grained representations.

\item \textbf{Dynamic Segmentation:} A learned boundary detector identifies semantic breakpoints by measuring local dissimilarity between adjacent token representations. Unlike LCM's fixed sentence boundaries, these boundaries emerge from the model's own latent space through end-to-end optimization.

\item \textbf{Concept-Level Reasoning:} Tokens within each segment are pooled into unified concept representations. A high-capacity transformer then performs deep reasoning exclusively on this compressed concept sequence---where the majority of computation occurs.

\item \textbf{Token-Level Decoding:} A decoder reconstructs token-level predictions by attending to the reasoned concepts via a causal cross-attention mechanism.
\end{enumerate}

This design explicitly decouples \emph{what to think about} (concept formation via learned boundaries) from \emph{how to think} (reasoning in compressed latent space), enabling the model to allocate computation adaptively based on semantic structure rather than surface token count. As our results demonstrate, this structural bias makes the model exceptionally proficient at handling high-information, low-predictability tokens that mark the beginning of new concepts.

Our main contributions are as follows:
\begin{itemize}
    \item \textbf{Concept-level latent reasoning with learned boundaries.}
    We propose DLCM, a hierarchical next-token prediction architecture that discovers variable-length semantic concepts from latent representations and performs deep computation in a compressed concept space.

    \item \textbf{Compression-aware scaling law for hierarchical LMs.}
    We derive a scaling law $L(N,D,R,P)$ that explicitly models the interaction between total parameters $N$, data $D$, compression ratio $R$, and concept-backbone allocation $P$, enabling principled selection of architecture under equal-FLOPs constraints. \cite{hoffmann2022chinchilla}
    
    \item \textbf{Decoupled $\mu$P for heterogeneous modules (why different LR / init variance).}
    We demonstrate that the Maximal Update Parametrization ($\mu$P) can be effectively adapted to our heterogeneous architecture to prevent training instability and ensure optimal performance at scale. Specifically, we identify that due to the decoupled widths of our model, the learning rates for the token-level components, concept backbone, and embeddings must be adjusted independently. Empirically, we confirm that the optimal effective learning rate for each component scales inversely with its specific width ($\eta \propto \text{width}^{-1}$), a finding that aligns with theoretical predictions for uniform models but is verified here in a non-uniform setting.

    \item \textbf{Compute redistribution yields reasoning gains at lower FLOPs.}
    With $R=4$, DLCM reduces FLOPs by up to 34\% while reallocating capacity into a larger reasoning backbone, improving average accuracy by 2.69\% on 12 zero-shot benchmarks, with the largest gains on reasoning-dominant tasks.
\end{itemize}
\section{Related Work}
\subsection{From Latent Reasoning to Concept-Level Language Modeling}

Recent research has explored performing reasoning at higher levels of abstraction than individual tokens, offering both computational efficiency and new modeling capabilities. \textbf{Latent reasoning} frameworks perform reasoning entirely within continuous hidden state spaces rather than through explicit token generation~\cite{hao2025traininglargelanguagemodels}. In the COCONUT framework, the model's hidden state from one reasoning step feeds directly into subsequent steps without generating intermediate tokens~\cite{hao2025traininglargelanguagemodels}. This approach offers significant computational advantages over methods like Chain-of-Thought prompting, which require generating hundreds of intermediate tokens. More importantly, continuous representations can encode multiple potential reasoning paths in superposition, enabling parallel exploration of the solution space~\cite{hao2025traininglargelanguagemodels}. However, this comes at the cost of interpretability and can struggle with tasks requiring precise symbolic manipulation.

Building upon latent reasoning principles, the \textbf{Large Concept Model (LCM)} framework operates at an intermediate level: reasoning on sentence-level "concepts" rather than tokens~\cite{lcm2024large}. LCMs use a three-stage pipeline: (1) a frozen encoder maps sentences to fixed-size embeddings in a semantic space (e.g., SONAR, supporting 200 languages), (2) a transformer performs autoregressive prediction in this concept space using diffusion or quantization techniques adapted from computer vision, and (3) a frozen decoder reconstructs token-level output~\cite{lcm2024large}. This approach combines key advantages: like latent reasoning, it operates in continuous space with substantial efficiency gains (10× sequence length reduction), but unlike pure latent reasoning, each concept remains interpretable as a decodable sentence~\cite{lcm2024large}. Remarkably, LCMs demonstrate zero-shot multilingual transfer—models trained only on English can generate in 200+ languages by leveraging the language-agnostic semantic space~\cite{lcm2024large}. However, the LCM framework faces significant limitations. First, it requires pre-training separate encoder and decoder models on massive multilingual data before the LCM itself can be trained, creating a scalability bottleneck~\cite{lcm2024large}. Second, and more fundamentally, the sentence-level granularity is a fixed human prior—the model must accept predetermined sentence boundaries rather than learning task-optimal segmentation. This rigidity prevents the model from adapting its conceptual granularity to different domains or tasks. Our DCLM architecture addresses both limitations through end-to-end training with dynamic, learnable boundary detection that discovers optimal chunking strategies directly from the data.

\subsection{Dynamic Compute Allocation in Language Models}

Standard LLMs allocate uniform computation to every token, ignoring that some tokens (e.g., predictable function words) require minimal processing while others (e.g., concept boundaries) demand more effort~\cite{kaplan2020scaling, hoffmann2022training}. Recent work explores adaptive allocation mechanisms. The Universal Transformer~\cite{dehghani2018universal} introduced recurrence in depth, applying the same transformation block repeatedly with a learned halting mechanism that determines when each position has been sufficiently refined. Mixture of Experts (MoE) models~\cite{shazeer2017outrageously, jiang2024mixtral} achieve conditional computation by routing each token to a subset of expert sub-networks. However, these approaches focus on parameter efficiency and scaling rather than fundamentally addressing the information density problem.

\textbf{H-NET}~\cite{hwang2025dynamicchunkingendtoendhierarchical} directly addresses adaptive allocation through learned boundary detection. The model predicts semantic boundaries by analyzing local patterns (e.g., similarity between consecutive hidden states), segments the sequence into variable-length chunks, and processes the compressed chunk representations hierarchically. Critically, boundary detection is differentiable and trained end-to-end, allowing task-appropriate chunking strategies to emerge~\cite{hwang2025dynamicchunkingendtoendhierarchical}. This yields substantial gains: learned boundaries align with linguistic structures even without supervision, and compression (4-8× reduction) translates to quadratic attention savings~\cite{hwang2025dynamicchunkingendtoendhierarchical}. The approach implicitly allocates more computation to high-information boundaries where new concepts begin—precisely where prediction is most difficult. However, H-NET's primary focus is on efficient representation through hierarchical bit-level modeling rather than token-level generation in state-of-the-art autoregressive LLMs~\cite{hwang2025dynamicchunkingendtoendhierarchical}. This leaves unaddressed the critical problem of computational waste in modern decoder-only language models, where every token—regardless of its predictability or information content—receives identical processing through the full model depth. Our DCLM architecture bridges this gap by adapting H-NET's dynamic boundary detection principles to the token-level generation paradigm of current LLMs~\cite{brown2020language, touvron2023llama}. By segmenting sequences into concept chunks and performing compressed reasoning before token-level decoding, DCLM enables adaptive computation allocation in the exact architectural context where it matters most: next-token prediction in large-scale autoregressive models.
\section{Methodology}

We now describe the technical details of \ModelName{}. The overall architecture is illustrated in Figure~\ref{fig:architecture}.

\subsection{Overview}

\ModelName{} processes a token sequence through four stages: (1) \textbf{Encoding} extracts fine-grained token representations; (2) \textbf{Dynamic Segmentation} identifies semantic boundaries and pools tokens into concepts; (3) \textbf{Concept-Level Reasoning} performs deep computation on the compressed sequence; and (4) \textbf{Token-Level Decoding} reconstructs predictions by attending to reasoned concepts. We formalize this as:
\begin{align}
    \mathbf{H} &= \mathcal{E}(\mathbf{x}) && \text{(Encoding)} \\
    \mathbf{C} &= \Phi(\mathbf{H}) && \text{(Segmentation \& Pooling)} \\
    \mathbf{Z} &= \mathcal{M}(\mathbf{C}) && \text{(Concept Reasoning)} \\
    \hat{\mathbf{y}} &= \mathcal{D}(\Psi(\mathbf{H}, \mathbf{Z})) && \text{(Decoding)}
\end{align}
where $\mathcal{E}$ is the encoder, $\Phi$ is the segmentation-pooling operation, $\mathcal{M}$ is the concept-level transformer, $\mathcal{D}$ is the decoder, and $\Psi$ is the cross-attention expansion defined in Eq.~\ref{eq:cross_attn}.

\subsection{Encoding}

The encoder $\mathcal{E}$ is a standard causal Transformer that processes raw tokens $\mathbf{x} = [x_1, \dots, x_L]$ to produce fine-grained representations $\mathbf{H} = [\mathbf{h}_1, \dots, \mathbf{h}_L] \in \mathbb{R}^{L \times d_{\text{token}}}$. These representations capture local contextual information and serve as the basis for both boundary detection and final token-level decoding.

\subsection{Dynamic Segmentation}

While boundary scores are learned end-to-end from latent representations, we intentionally decouple the discrete segmentation decision from the language modeling loss to avoid optimization interference.
This design trades full end-to-end discreteness for training stability and controllable compression, which we find essential at scale.

\subsubsection{Boundary Detection}

Our key hypothesis is that transitions between distinct concepts are marked by significant shifts in the latent feature space. We detect these ``semantic breaks'' by measuring local dissimilarity between adjacent tokens.

Given encoder outputs $\mathbf{H}$, we project each token into a query-key space of dimension $d_{\text{scan}}$:
\begin{equation}
    \mathbf{q}_t = \mathbf{W}_q \mathbf{h}_t, \quad \mathbf{k}_t = \mathbf{W}_k \mathbf{h}_t
\end{equation}
The boundary probability $p_t \in [0,1]$ is computed as the normalized dissimilarity:
\begin{equation}
    p_t = \frac{1 - \cos(\mathbf{q}_{t-1}, \mathbf{k}_t)}{2} = \frac{1}{2} \left( 1 - \frac{\mathbf{q}_{t-1}^{\top}\mathbf{k}_t}{\|\mathbf{q}_{t-1}\|_2 \|\mathbf{k}_t\|_2} \right)
    \label{eq:boundary}
\end{equation}
We enforce $p_1 = 1$ so that the first token always starts a new concept.

\paragraph{Discrete Sampling.}
While $p_t$ is continuous, downstream processing requires discrete segment assignments. We adopt different strategies for training and inference:
\begin{itemize}
    \item \textbf{Training:} We sharpen probabilities by temperature $\alpha$, then sample $b_t \sim \text{Bernoulli}(p_t^{\text{sharp}})$ to encourage exploration.
    \item \textbf{Inference:} We use a \textbf{hard thresholding} rule: $b_t = [\, p_t \ge 0.5 \,]$.
\end{itemize}

\subsubsection{Concept Formation via Pooling}

Given boundary indicators $\mathbf{b} = [b_1, \dots, b_L]$, we partition tokens into $M$ contiguous segments $S_1, \dots, S_M$. Each segment is compressed into a single concept representation via mean pooling, followed by a projection to the concept dimension $d_{\text{concept}}$:
\begin{equation}
    \mathbf{c}_k^{\text{raw}} = \frac{1}{|S_k|} \sum_{t \in S_k} \mathbf{h}_t, \quad \mathbf{c}_k = \mathbf{W}_{\text{up}}\mathbf{c}_k^{\text{raw}}
\end{equation}
where $\mathbf{W}_{\text{up}} \in \mathbb{R}^{d_{\text{concept}} \times d_{\text{token}}}$ aligns the feature space. The resulting concept sequence $\mathbf{C} = [\mathbf{c}_1, \dots, \mathbf{c}_M]$ has length $M \ll L$.

\subsubsection{Adaptive Compression via Global Load Balancing}
\label{sec:adaptive_compression}

Similar to H-Net~\cite{hwang2025dynamicchunkingendtoendhierarchical}, natural language exhibits varying information density. To enable content-adaptive compression while maintaining a target ratio $R$ (e.g., $R=4$ means 4 tokens per concept on average), we impose constraints at the \emph{global batch level} rather than per-sequence.

Let $\mathcal{T}$ denote all tokens across the distributed batch. We track:
\begin{align}
    G_{\text{global}} &= \frac{1}{|\mathcal{T}|} \sum_{(i,t) \in \mathcal{T}} p_{i,t} && \text{(expected boundary rate)} \\
    F_{\text{global}} &= \frac{1}{|\mathcal{T}|} \sum_{(i,t) \in \mathcal{T}} b_{i,t} && \text{(actual boundary rate)}
\end{align}
These statistics are synchronized across ranks via \texttt{AllReduce}. We optimize an auxiliary loss:
\begin{equation}
    \mathcal{L}_{\text{aux}} = \frac{R}{R-1} \left[ (R-1) \cdot F_{\text{global}} \cdot G_{\text{global}} + (1 - F_{\text{global}}) \cdot (1 - G_{\text{global}}) \right] - 1
\end{equation}
This encourages the global compression rate to converge to $1/R$ while allowing local fluctuation. We refer to this globally regularized segmentation mechanism as the \textit{Global Parser}, which serves as a critical component for enabling content-adaptive granularity.

\subsection{Concept-Level Reasoning}
\label{sec:backbone}

The concept-level transformer $\mathcal{M}$ is the computational core of \ModelName{}. Operating on the compressed sequence $\mathbf{C} \in \mathbb{R}^{M \times d_{\text{concept}}}$, it performs deep reasoning with significantly reduced attention complexity.

$\mathcal{M}$ is a standard causal Transformer with $L_{\text{concept}}$ layers. Because concepts represent semantic units rather than surface tokens, this module can focus on high-level reasoning without being distracted by low-level token prediction. The output $\mathbf{Z} = \mathcal{M}(\mathbf{C})$ contains enriched concept representations.

\subsection{Token-Level Decoding}
\label{sec:decoding}

The decoder reconstructs token-level predictions by attending to the reasoned concepts. This involves two components: concept smoothing and causal cross-attention.

\subsubsection{Concept Smoothing}

Hard pooling can introduce discretization artifacts at segment boundaries. We apply a lightweight smoothing module $\mathcal{S}$ to integrate adjacent concepts:
\begin{equation}
    \tilde{\mathbf{Z}} = \mathcal{S}(\mathbf{Z})
\end{equation}

\subsubsection{Causal Cross-Attention}

The decoder $\mathcal{D}$ generates token predictions by querying the smoothed concepts. Crucially, we enforce causality so token $t$ can only attend to concepts formed up to index $j(t) = \sum_{i=1}^t b_i$.

To handle the heterogeneous architecture ($d_{\text{token}} \neq d_{\text{concept}}$), the cross-attention mechanism projects queries from the encoder space and keys/values from the concept space into a common head dimension $d_{\text{head}}$:
\begin{align}
    \mathbf{Q} &= \mathbf{H}\mathbf{W}_Q, \quad \text{where } \mathbf{W}_Q \in \mathbb{R}^{d_{\text{token}} \times d_{\text{head}}} \\
    \mathbf{K} &= \tilde{\mathbf{Z}}\mathbf{W}_K, \quad \mathbf{V} = \tilde{\mathbf{Z}}\mathbf{W}_V, \quad \text{where } \mathbf{W}_{K,V} \in \mathbb{R}^{d_{\text{concept}} \times d_{\text{head}}}
\end{align}

The attention output is computed as:
\begin{equation}
    \Psi(\mathbf{H}, \mathbf{Z}) = \text{Softmax}\left( \frac{\mathbf{Q}\mathbf{K}^\top}{\sqrt{d_{\text{head}}}} + \mathbf{M} \right) \mathbf{V}\mathbf{W}_O + \mathbf{H}
    \label{eq:cross_attn}
\end{equation}
where $\mathbf{W}_O \in \mathbb{R}^{d_{\text{head}} \times d_{\text{token}}}$ projects the result back to the token dimension for the residual connection.

\subsection{Training Objective}

The total loss combines next-token prediction with adaptive compression:
\begin{equation}
    \mathcal{L} = \mathcal{L}_{\text{CE}} + \lambda \mathcal{L}_{\text{aux}}
\end{equation}
where $\mathcal{L}_{\text{CE}}$ is cross-entropy on output tokens and $\mathcal{L}_{\text{aux}}$ is the load-balancing loss.

\section{Implementation Details}

\paragraph{Packed Sequence Training.}
We adopt the Variable Length (VarLen) approach from FlashAttention~\cite{dao2022flashattention} to ensure global compression statistics are computed over diverse tokens.

\paragraph{QK Normalization.}
Bridging token-level and concept-level representations with different statistical properties can cause training instability. Following~\cite{henry2020query, dehghani2023scaling}, we apply RMSNorm to queries and keys before attention:
\begin{equation}
    \mathbf{Q}' = \text{RMSNorm}(\mathbf{Q}), \quad \mathbf{K}' = \text{RMSNorm}(\mathbf{K})
\end{equation}

\subsection{Efficient Cross-Attention via Concept Replication}

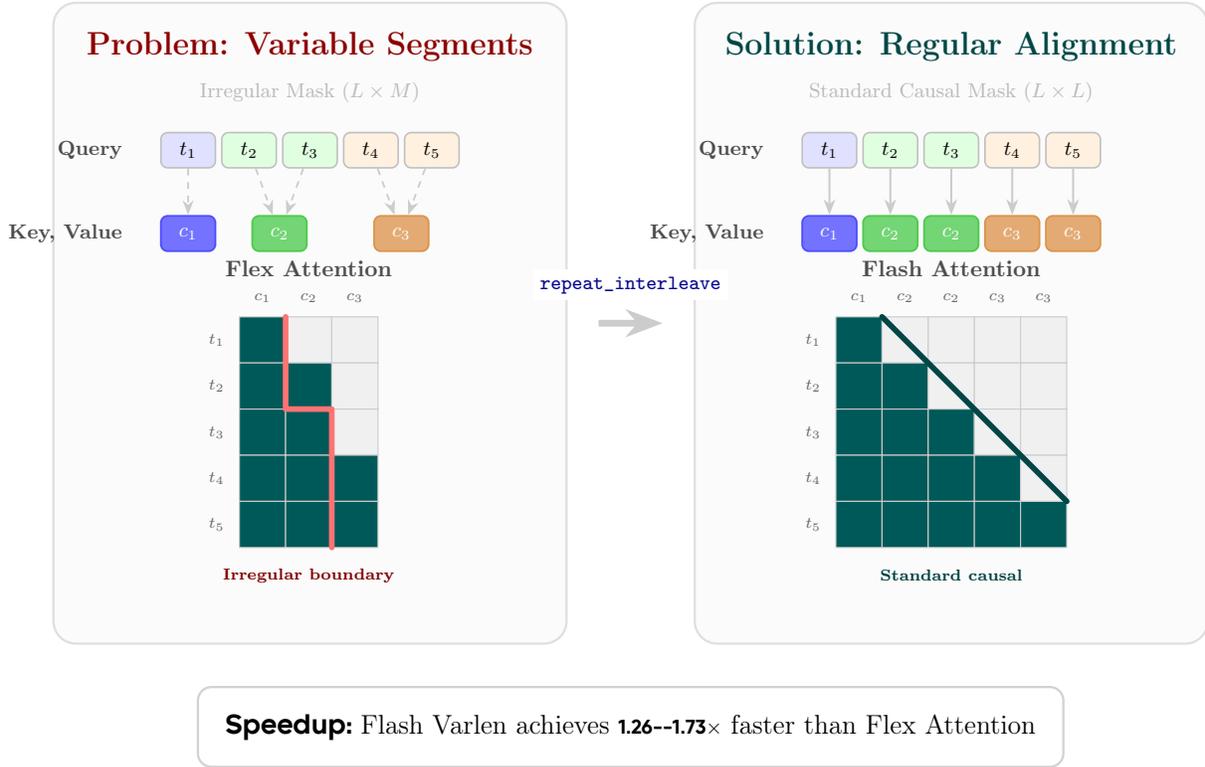
\begin{figure*}[t]
\centering
\resizebox{0.98\textwidth}{!}{
\begin{tikzpicture}[
    >=Stealth,
    font=\sffamily,
    panel style/.style={
        draw=gray!25, fill=gray!3, rounded corners=10pt, line width=1pt
    },
    title prob/.style={text=red!55!black, font=\Large\bfseries},
    title sol/.style={text=teal!55!black, font=\Large\bfseries},
    token box/.style={
        draw=gray!50, fill=white, rounded corners=3pt, line width=0.7pt,
        minimum width=0.85cm, minimum height=0.55cm, font=\bfseries\small, align=center
    },
    concept box/.style={
        draw=#1!70, fill=#1!55, rounded corners=3pt, line width=0.8pt,
        minimum width=0.85cm, minimum height=0.55cm, font=\bfseries\small, align=center, text=white
    },
    row label/.style={font=\small\bfseries, text=black!70, anchor=east},
    grid line/.style={draw=gray!40, line width=0.5pt},
    active cell/.style={fill=teal!70!black},
    masked cell/.style={fill=gray!12},
    highlight line/.style={line width=2.5pt, line cap=round, line join=round}
]

\def\leftcenter{4.0}
\def\rightcenter{14.0}
\def\panelwidth{8.0}
\def\panelheight{10.0}

\node[panel style, minimum width=\panelwidth cm, minimum height=\panelheight cm] (left_panel) at (\leftcenter, -5) {};
\node[panel style, minimum width=\panelwidth cm, minimum height=\panelheight cm] (right_panel) at (\rightcenter, -5) {};

\pgfmathsetmacro{\arrowstart}{\leftcenter + \panelwidth/2 + 0.5}
\pgfmathsetmacro{\arrowend}{\rightcenter - \panelwidth/2 - 0.5}
\pgfmathsetmacro{\arrowmid}{(\arrowstart+\arrowend)/2}

\draw[->, line width=3pt, gray!40] (\arrowstart, -5.0) -- (\arrowend, -5.0);
\node[font=\ttfamily\bfseries\small, text=blue!55!black, fill=white, inner sep=3pt] at (\arrowmid, -4.4) {repeat\_interleave};

\node[title prob, anchor=north, yshift=-10pt] at (left_panel.north) {Problem: Variable Segments};
\node[text=gray!55, font=\small, anchor=north, yshift=-32pt] at (left_panel.north) {Irregular Mask ($L \times M$)};

\begin{scope}[shift={(\leftcenter, -2.3)}]
    \def\tokenspacing{0.95}
    \def\labeloffset{-2.8}
    
    \node[row label] at (\labeloffset, 0) {Query};
    \foreach \i in {1,...,5} {
        \pgfmathsetmacro{\xpos}{(\i-3)*\tokenspacing}
        \pgfmathtruncatemacro{\idx}{\i}
        \ifnum\i=1
            \node[token box, fill=blue!12] (t\i) at (\xpos, 0) {$t_\idx$};
        \else\ifnum\i<4
            \node[token box, fill=green!12] (t\i) at (\xpos, 0) {$t_\idx$};
        \else
            \node[token box, fill=orange!12] (t\i) at (\xpos, 0) {$t_\idx$};
        \fi\fi
    }

    \node[row label] at (\labeloffset, -1.3) {Key, Value};
    \node[concept box=blue] (c1) at (-2*\tokenspacing, -1.3) {$c_1$};
    \node[concept box=green!70!black] (c2) at (-0.5*\tokenspacing, -1.3) {$c_2$};
    \node[concept box=orange!80!black] (c3) at (1.5*\tokenspacing, -1.3) {$c_3$};

    \draw[->, gray!40, dashed, line width=0.8pt] (t1) -- (c1);
    \draw[->, gray!40, dashed, line width=0.8pt] (t2) -- (c2);
    \draw[->, gray!40, dashed, line width=0.8pt] (t3) -- (c2);
    \draw[->, gray!40, dashed, line width=0.8pt] (t4) -- (c3);
    \draw[->, gray!40, dashed, line width=0.8pt] (t5) -- (c3);
\end{scope}

\begin{scope}[shift={(\leftcenter - 1.1, -8.5)}, scale=0.72]
    \node[font=\bfseries\color{black!70}, anchor=south] at (1.5, 5.7) {Flex Attention};
    
    \foreach \r/\label in {1/$t_1$, 2/$t_2$, 3/$t_3$, 4/$t_4$, 5/$t_5$} {
        \node[font=\scriptsize\color{black!60}, anchor=east] at (-0.15, 5.5-\r) {\label};
    }
    \foreach \c/\label in {1/$c_1$, 2/$c_2$, 3/$c_3$} {
        \node[font=\scriptsize\color{black!60}, anchor=south] at (\c-0.5, 5.15) {\label};
    }
    
    \foreach \r in {1,...,5} \foreach \c in {1,2,3} {
        \pgfmathtruncatemacro{\on}{
            (\r==1 && \c==1) || (\r==2 && \c<=2) || (\r==3 && \c<=2) || (\r>=4 && \c<=3) ? 1 : 0
        }
        \ifnum\on=1 \fill[active cell] (\c-1, 5-\r) rectangle +(1,1);
        \else \fill[masked cell] (\c-1, 5-\r) rectangle +(1,1); \fi
        \draw[grid line] (\c-1, 5-\r) rectangle +(1,1);
    }
    \draw[red!55, highlight line] (1,5) -- (1,3) -- (2,3) -- (2,0);
    
    \node[font=\scriptsize\bfseries\color{red!50!black}] at (1.5, -0.6) {Irregular boundary};
\end{scope}

\node[title sol, anchor=north, yshift=-10pt] at (right_panel.north) {Solution: Regular Alignment};
\node[text=gray!55, font=\small, anchor=north, yshift=-32pt] at (right_panel.north) {Standard Causal Mask ($L \times L$)};

\begin{scope}[shift={(\rightcenter, -2.3)}]
    \def\tokenspacing{0.95}
    \def\labeloffset{-2.8}
    
    \node[row label] at (\labeloffset, 0) {Query};
    \foreach \i in {1,...,5} {
        \pgfmathsetmacro{\xpos}{(\i-3)*\tokenspacing}
        \pgfmathtruncatemacro{\idx}{\i}
        \ifnum\i=1
            \node[token box, fill=blue!12] (rt\i) at (\xpos, 0) {$t_\idx$};
        \else\ifnum\i<4
            \node[token box, fill=green!12] (rt\i) at (\xpos, 0) {$t_\idx$};
        \else
            \node[token box, fill=orange!12] (rt\i) at (\xpos, 0) {$t_\idx$};
        \fi\fi
    }

    \node[row label] at (\labeloffset, -1.3) {Key, Value};
    \foreach \i in {1,...,5} {
        \pgfmathsetmacro{\xpos}{(\i-3)*\tokenspacing}
        \ifnum\i=1
            \node[concept box=blue] (rc\i) at (\xpos, -1.3) {$c_1$};
        \else\ifnum\i<4
            \node[concept box=green!70!black] (rc\i) at (\xpos, -1.3) {$c_2$};
        \else
            \node[concept box=orange!80!black] (rc\i) at (\xpos, -1.3) {$c_3$};
        \fi\fi
    }

    \foreach \i in {1,...,5} {
        \draw[->, gray!40, line width=0.9pt] (rt\i) -- (rc\i);
    }
\end{scope}

\begin{scope}[shift={(\rightcenter - 1.8, -8.5)}, scale=0.72]
    \node[font=\bfseries\color{black!70}, anchor=south] at (2.5, 5.7) {Flash Attention};
    
    \foreach \r/\label in {1/$t_1$, 2/$t_2$, 3/$t_3$, 4/$t_4$, 5/$t_5$} {
        \node[font=\scriptsize\color{black!60}, anchor=east] at (-0.15, 5.5-\r) {\label};
    }
    \foreach \c/\label in {1/$c_1$, 2/$c_2$, 3/$c_2$, 4/$c_3$, 5/$c_3$} {
        \node[font=\scriptsize\color{black!60}, anchor=south] at (\c-0.5, 5.15) {\label};
    }
    
    \foreach \r in {1,...,5} \foreach \c in {1,...,5} {
        \pgfmathtruncatemacro{\on}{\c <= \r ? 1 : 0}
        \ifnum\on=1 \fill[active cell] (\c-1, 5-\r) rectangle +(1,1);
        \else \fill[masked cell] (\c-1, 5-\r) rectangle +(1,1); \fi
        \draw[grid line] (\c-1, 5-\r) rectangle +(1,1);
    }
    \draw[teal!55!black, highlight line] (1,5) -- (2,4) -- (3,3) -- (4,2) -- (5,1);
    
    \node[font=\scriptsize\bfseries\color{teal!50!black}] at (2.5, -0.6) {Standard causal};
\end{scope}

\node[draw=gray!35, fill=white, rounded corners=6pt, line width=1pt, inner sep=12pt] at ({(\leftcenter+\rightcenter)/2}, -11.3) {
    \large\bfseries Speedup: \normalfont\large Flash Varlen achieves \textbf{1.26--1.73$\times$} faster than Flex Attention
};

\end{tikzpicture}
}
\caption{\textbf{Cross-Attention Optimization via Concept Replication.} 
\textit{Left:} The decoder's cross-attention creates an irregular $L \times M$ mask due to variable token-to-concept mappings. 
\textit{Right:} By replicating concepts via \texttt{repeat\_interleave} to match token positions, we obtain a standard $L \times L$ causal mask, enabling optimized Flash Attention kernels.}
\label{fig:cross_attn_optimization}
\end{figure*}

The decoder's cross-attention mechanism (Section~\ref{sec:decoding}) presents a significant implementation challenge. Mathematically, tokens must attend to concepts with variable-length mappings ($L \times M$), creating irregular attention patterns. As illustrated in Figure~\ref{fig:cross_attn_optimization}, when tokens $\{t_1\}$ belong to concept $c_1$, and tokens $\{t_2, t_3\}$ belong to $c_2$, the resulting attention mask effectively has a "ragged" boundary.

Implementing this directly with \textbf{Flex Attention} incurs significant overhead from dynamic mask generation and irregular memory access patterns. To address this, we adopt a \textbf{concept replication strategy} to bridge the gap between the theoretical $L \times M$ formulation and hardware-friendly $L \times L$ kernels. Analogous to Grouped Query Attention (GQA), we expand concepts via \texttt{repeat\_interleave} to match token positions.

Specifically, for each token $t_i$ belonging to concept $c_j$, we replicate the concept feature $c_j$ at position $i$ in the key/value sequence:
\begin{equation}
\tilde{\mathbf{K}} = \texttt{repeat\_interleave}(\mathbf{K}, \text{segment\_lengths}), \quad \tilde{\mathbf{V}} = \texttt{repeat\_interleave}(\mathbf{V}, \text{segment\_lengths})
\end{equation}
This transformation aligns the Key/Value length with the Query length ($L$), enabling the use of \textbf{Flash Attention with Variable Length (Varlen)}. This allows us to leverage highly optimized CUDA kernels designed for standard causal masking, treating the problem as a specialized form of self-attention where keys and values are locally constant within each concept segment.

\subsection{Performance Benchmarks}

We benchmark the efficiency of our concept replication strategy against Flex Attention using independent kernel profiling. Note that the hidden sizes listed here (1024, 2048, 4096) are standard benchmarking dimensions and do not necessarily match the specific architectural dimensions ($d_{\text{token}}$, $d_{\text{concept}}$) of \ModelName{}, as the primary goal is to demonstrate algorithmic scalability.

Table~\ref{tab:perf_comparison} provides detailed performance . To more intuitively analyze the performance trends, we have plotted the speedup ($T_{\text{flex}} / T_{\text{flash}}$) in Figure~\ref{fig:speedup_plot}.

\subsection{Key Observations and Analysis}

\begin{figure}[h]
\centering
\includegraphics[width=0.9\textwidth]{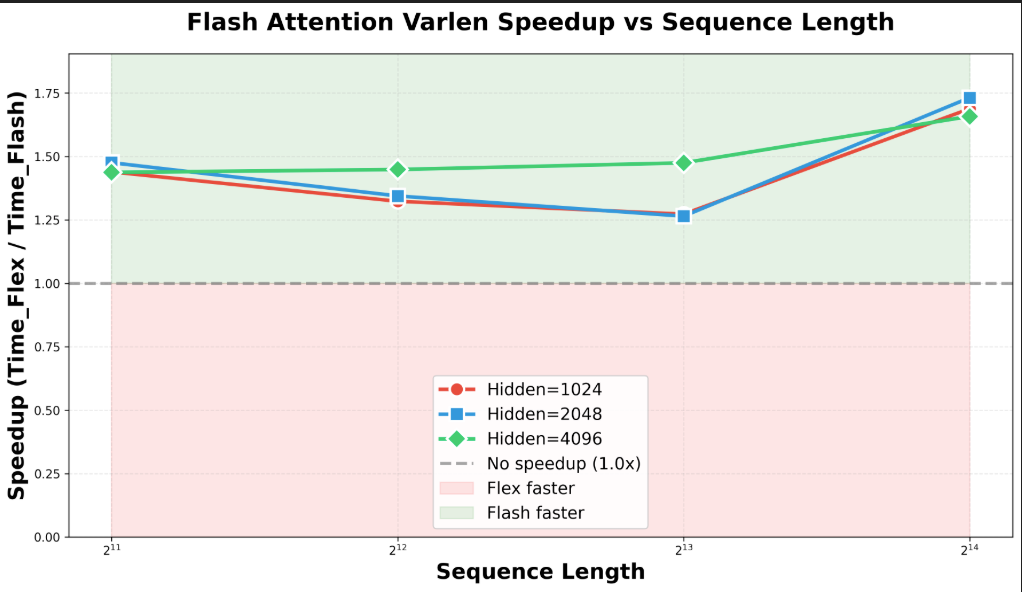}
\caption{Flash Attention Varlen speedup ($T_{\text{flex}} / T_{\text{flash}}$) vs. Sequence Length. The plot visualizes the data from Table~\ref{tab:perf_comparison}, highlighting the performance trend across different scales and hidden sizes.}
\label{fig:speedup_plot}
\end{figure}

We can draw three key conclusions from these results:

\begin{itemize}
    \item \textbf{Consistent Performance Advantage:} In all tested configurations, Flash Attention Varlen with concept replication significantly outperforms Flex Attention. The speedup ranges from \textbf{1.26$\times$ to 1.73$\times$}, validating the efficacy of the ``memory-for-computation'' trade-off.

    \item \textbf{Insensitivity to Hidden Size:} As shown in Figure~\ref{fig:speedup_plot}, the three lines representing different hidden sizes (1024, 2048, and 4096) are nearly identical. This indicates that the performance bottleneck is dominated by the memory access patterns of the attention mechanism, not the computational complexity of the hidden dimension. Flash Varlen's optimized, regular memory access pattern remains stable across various model widths.

    \item \textbf{Superior Scalability with Sequence Length:} The most critical finding is that Flash Varlen's performance advantage \textbf{scales with increasing sequence length}. At a 2K sequence length, the average speedup is $\sim$1.44$\times$. When the sequence length increases 8-fold to 16K, the average speedup climbs to $\sim$1.70$\times$, peaking at \textbf{1.73$\times$} for a hidden size of 2048.
\end{itemize}

This scalability trend strongly suggests that the overhead from Flex Attention's dynamic mask generation and irregular memory access patterns grows faster than the computational cost. Conversely, despite its increased memory footprint for the K/V cache, Flash Varlen's highly optimized kernel and regular causal access pattern prove far more efficient, especially at longer sequences.

\section{Data}
To ensure experimental reproducibility, we build our corpus entirely from open-source data and tokenize it using the DeepSeek-v3 \cite{liu2024deepseek} tokenizer. Our corpus spans multiple domains, including web text (English and Chinese), mathematics, and code, forming a comprehensive foundation for core language understanding across linguistic, factual, and reasoning abilities.

The composition of our pretraining data is intentionally designed to serve two critical objectives. First, it balances breadth and specialization: web text provides broad natural language coverage, while mathematics and code enhance structured reasoning. Second, and more importantly for our architecture, this diversity is essential for learning robust dynamic segmentation. By exposing the model to domains with drastically different information densities (e.g., highly structured code syntax vs. verbose natural language prose), we force the learned boundary predictor to discover content-adaptive segmentation strategies that generalize across diverse tasks. English and Chinese web text are weighted more heavily to ensure multilingual alignment, while specialized datasets like MegaMath-Web and OpenCoder-Pretrain are included to fine-tune the model's handling of high-entropy transitions.

To demonstrate the architectural benefits of \ModelName{} rather than gains from data curation, we do not apply aggressive filtering; instead, we use data whose quality aligns with standard open-source corpora. \Cref{tab:training_data_details} summarizes the statistics.

\begin{table}[ht]
\centering
\small
\caption{\textbf{Statistics of the pretraining data.}}
\label{tab:training_data_details}
\begin{tabular}{l|cc}
\toprule
\textbf{Data Source} & \textbf{Ratio} & \textbf{Tokens (B)} \\
\hline
Nemotron-CC \cite{su2024nemotron} (English Web) & 50\% & 500 \\
MAP-CC \cite{du2024chinese} (Chinese Web) & 25\% & 250 \\
OpenCoder-Pretrain \cite{Huang2024OpenCoderTO} & 15\% & 150 \\
MegaMath-Web \cite{zhou2025megamath} & 10\% & 100 \\
\hline
\textbf{Total} & \textbf{100\%} & \textbf{1,000} \\
\bottomrule
\end{tabular}
\end{table}

\section{Scaling Laws for \ModelName{}}

To determine the optimal architecture and hyperparameters for \ModelName{}, we conduct a comprehensive exploration using scaling laws. We first introduce a decoupled optimization strategy to handle the heterogeneous nature of our architecture, followed by the mathematical formulation of our scaling objectives.

\subsection{Decoupled $\mu$P for Heterogeneous Architectures}

\subsubsection{Formulation of $\mu$P}

To ensure consistent feature learning dynamics across varying scales and compression rates, we adopt the Maximal Update Parametrization ($\mu$P). Unlike standard transformers with uniform width, our architecture requires decoupled scaling for the token-level components ($\mathcal{E}, \mathcal{D}$) with width $d_{\text{token}}$ and the concept-level backbone ($\mathcal{M}$) with width $d_{\text{concept}}$.

We define distinct width multipliers relative to a base width $d_{\text{base}}$:
\begin{equation}
    s_{\text{token}} = \frac{d_{\text{token}}}{d_{\text{base}}}, \quad s_{\text{concept}} = \frac{d_{\text{concept}}}{d_{\text{base}}}
\end{equation}

Following standard $\mu$P practice, we adjust initialization variances and optimization hyperparameters separately for each component group:

\begin{itemize}
    \item \textbf{Initialization:} All hidden linear weights $W \in \mathbb{R}^{d_{\text{out}} \times d_{\text{in}}}$ are initialized with variance $\sigma_{\text{base}}^2 \cdot s^{-1}$, where $s \in \{s_{\text{token}}, s_{\text{concept}}\}$ corresponds to the layer's width. Embedding weights use fixed $\sigma_{\text{base}}^2$.
    
    \item \textbf{Learning Rates:} To stabilize feature learning, hidden layer learning rates are scaled inversely to width:
    \begin{align}
        \eta_{\mathcal{E}, \mathcal{D}} &= \eta^{\text{base}}_{\text{token}} \cdot s_{\text{token}}^{-1} \\
        \eta_{\mathcal{M}} &= \eta^{\text{base}}_{\text{concept}} \cdot s_{\text{concept}}^{-1}
    \end{align}
    Biases and embedding weights retain the fixed learning rate $\eta^{\text{base}}_{\text{others}}$.

    \item \textbf{Output Scaling:} To ensure the logits remain $O(1)$ effectively, the final decoder projection $W_{\text{unemb}}$ is scaled during the forward pass:
    \begin{equation}
        \text{logits} = \frac{1}{s_{\text{token}}} \cdot (\mathbf{h}_{\text{final}} W_{\text{unemb}}^\top)
    \end{equation}

    \item \textbf{Optimizer Stability:} The AdamW $\epsilon$ parameter for each layer is scaled by $s^{-1}$, matching the respective component width.
\end{itemize}

\subsubsection{Hyperparameter Tuning and Verification}
\begin{figure}[t]
    \centering
    \includegraphics[width=0.47\textwidth]{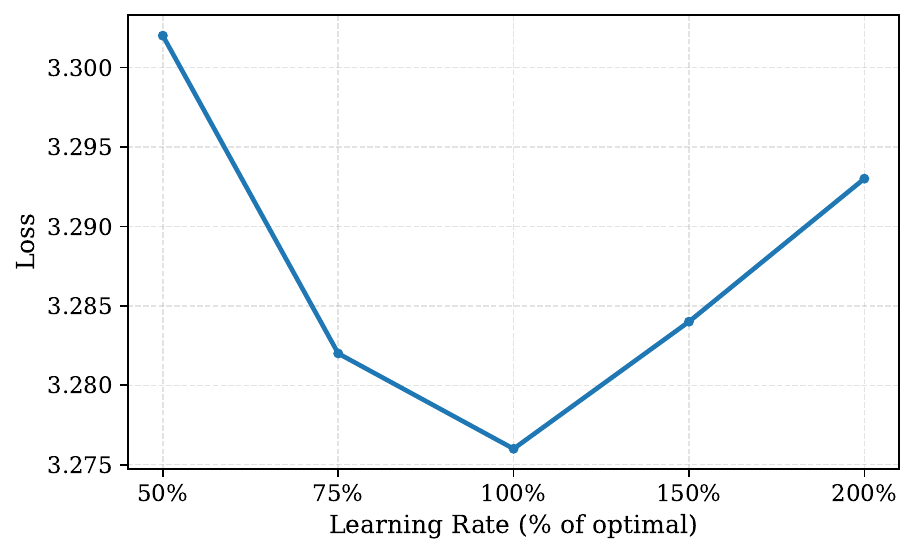}
    \hfill
    \includegraphics[width=0.47\textwidth]{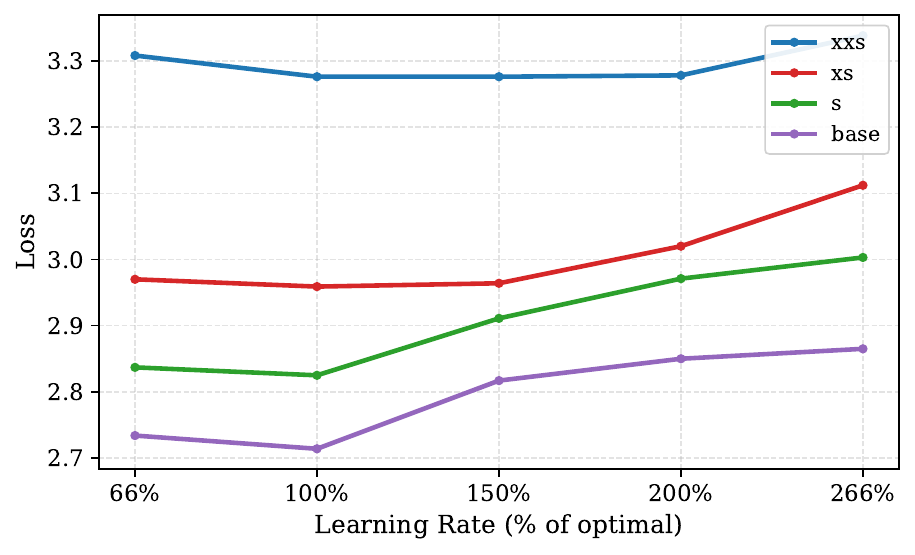}
    \caption{
    \textbf{Hyperparameter tuning and transfer under $\mu$P.}
    \textbf{Left:} We sweep $\eta^{\text{base}}_{\text{concept}}$ while fixing $\eta^{\text{base}}_{\text{token}}$ at its $\mu$P-predicted value on the 87M proxy model. The loss curve shows a well-defined minimum near $100\%$.
    \textbf{Right:} We jointly scale both base learning rates by the same factor and observe consistent minima across model sizes (87M–834M), validating the zero-shot transferability of $\mu$P-derived hyperparameters.
    }
    \label{mup_exp}
\end{figure}

Following the protocol proposed by \citet{yang2022tensor}, we adopted a two-stage strategy: tuning hyperparameters on a small proxy model and verifying their transferability on larger scales.

\paragraph{Proxy Model Tuning.} We performed coordinate descent on the base learning rates using a proxy model with 87M parameters. For each hyperparameter group, we iteratively swept over a multiplicative grid of $\{0.5, 0.75, 1.5, 2.0\}$ relative to the current best value until the validation loss stabilized. Empirically, we observed that the optimal base learning rates for the token and concept components were approximately equal ($\eta^{\text{base}}_{\text{concept}} \approx \eta^{\text{base}}_{\text{token}}$). We show the result of tuning $\eta^{\text{base}}_{\text{concept}}$ while keeping $\eta^{\text{base}}_{\text{token}}$ in Figure~\ref{mup_exp}, which means the actual learning rates depend on the ratio of the widths between the token and concept components in this unequal-width model. This consistency suggests that the explicit width-dependent scaling factors defined previously successfully account for the structural differences between components, stabilizing the effective learning rates across different widths.

\paragraph{Transfer Verification.} To validate the zero-shot hyperparameter transfer, we trained larger models (274M, 468M, 834M parameters) using the optimal $\eta^{\text{base}}$ values derived from the 87M proxy. Then we perturbed the predicted learning rates by simultaneously scaling $\eta^{\text{base}}_{\text{token}}$ and $\eta^{\text{base}}_{\text{concept}}$. As shown in Figure~\ref{mup_exp}, deviating from the $\mu$P-predicted learning rates resulted in degraded performance, confirming that the optimal hyperparameters found on the proxy model transfer effectively to larger scales without further tuning. Overall, this confirms that $\mu$P effectively stabilizes training for our unequal-width architecture, provided that the learning rates for the token and concept components are decoupled and scaled inversely to their respective widths. This finding extends standard scaling laws to heterogeneous designs, ensuring consistent optimality across scales.

\subsection{Scaling Law Formulation}

\subsubsection{Experimental Setup}
To validate our scaling hypotheses, we constructed a grid of models by varying the concept-layer parameter ratio $P \in \{30\%, 50\%, 70\%\}$ and the compression ratio $R \in \{2, 4, 8\}$. Models were trained on a budget of $200$B tokens, resulting in three primary model scales for analysis: \textbf{Small} (274M), \textbf{Medium} (468M), and \textbf{Large} (833M). All scaling exponents ($\delta_1, \delta_2, \gamma$) are \emph{shared globally} across all model scales and compression ratios, and are fitted \emph{once} using the joint training trajectories.
Only scale-independent offset terms are allowed to vary across configurations.
This design constrains the degrees of freedom of the model and avoids post-hoc overfitting to individual scales.

\subsubsection{Mathematical Formulation}
We extend the Chinchilla scaling framework~\cite{hoffmann2022training} by introducing (i) compression-aware behaviour and (ii) architectural decomposition. The resulting loss law is:

\begin{equation}
\label{eq:scaling_law}
L(N, D, R, P) 
= E_0
+ \frac{A_{\text{token}}}
       {(N(1-P) + t_{\text{token}})^{\delta_1}}
+ \frac{A_{\text{concept}} \, R^{\gamma}}
       {(NP + t_{\text{concept}})^{\delta_2}}
+ \frac{A_{\text{data}}}
       {(D + t_{\text{data}})^{\alpha}}.
\end{equation}

where $N$ is total parameters, $D$ is dataset size, $R$ is compression ratio, $P$ is concept-layer parameter ratio, and $E_0$ is the irreducible loss floor. This formulation disentangles token-processing efficiency, concept-processing efficiency (controlled by exponent $\gamma$), and data scaling.

\subsubsection{Decay-Phase Power Law}
Since our training protocol involves Weight-Sharing-and-Decay (WSD), we explicitly model the late-stage regime. We fit a simplified decay law to the fractional loss reduction $\Delta_{\text{decay}}$ in the 90\%–99\% token window:
\begin{equation}
\Delta_{\text{decay}} = k \, L_{\text{stable}}^{\,a} R^{\,b} N^{\,c}
\end{equation}
Obtained via log-linear regression, this model achieves $R^2 = 0.93$, accurately predicting late-stage loss drops across all scales.

\begin{figure}[t]
  \centering
  \includegraphics[width=0.9\linewidth]{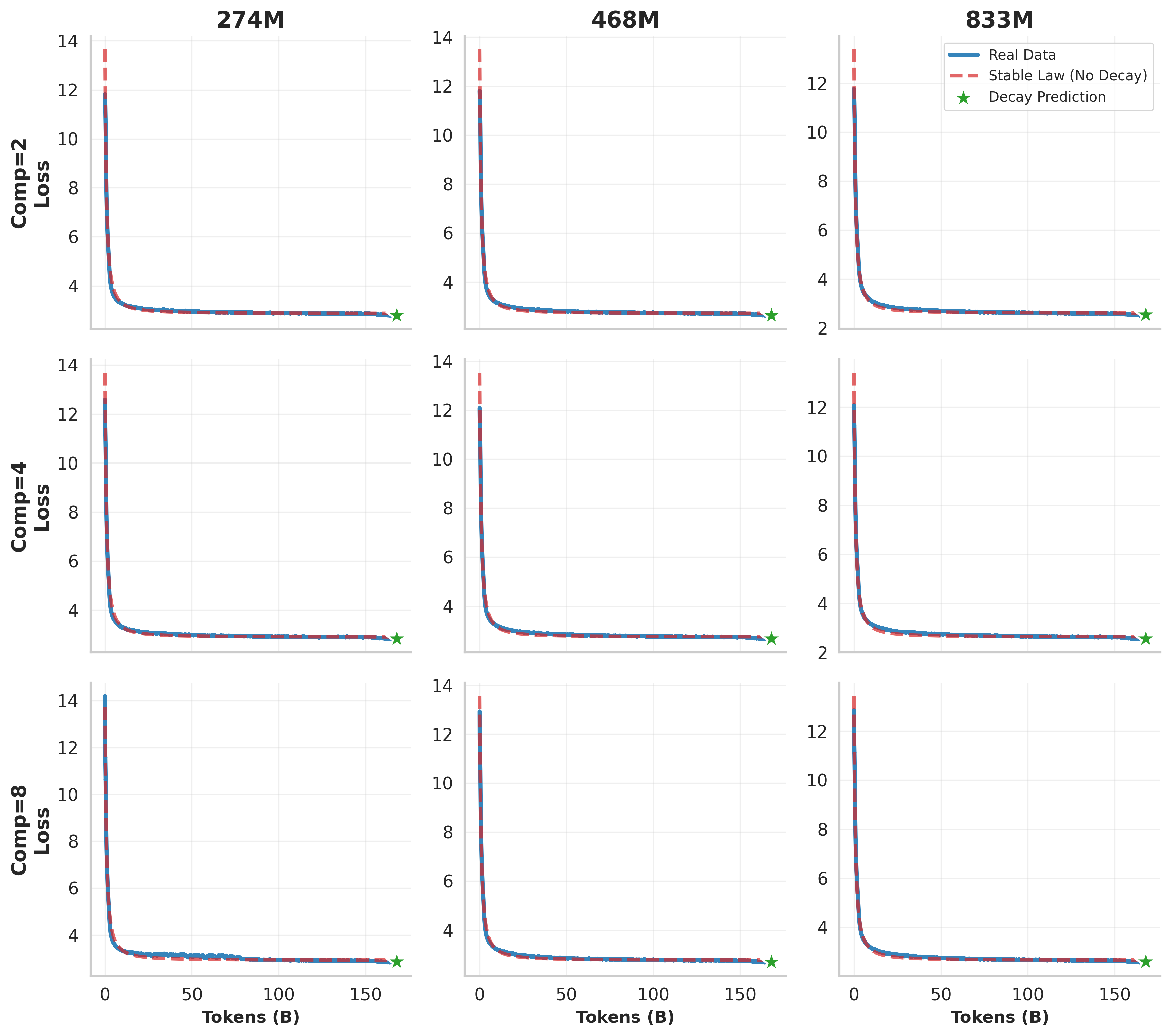}
  \caption{\textbf{Full training trajectory fit.} Comparison between predicted loss (Equation~\ref{eq:scaling_law}) and empirical loss across model sizes (274M--833M), compression factors $R \in \{2,4,8\}$, and training budgets. The joint fit achieves $R^2 > 0.98$.}
  \label{fig:full_training_fit}
\end{figure}

\begin{figure}[t]
  \centering
  \includegraphics[width=0.9\linewidth]{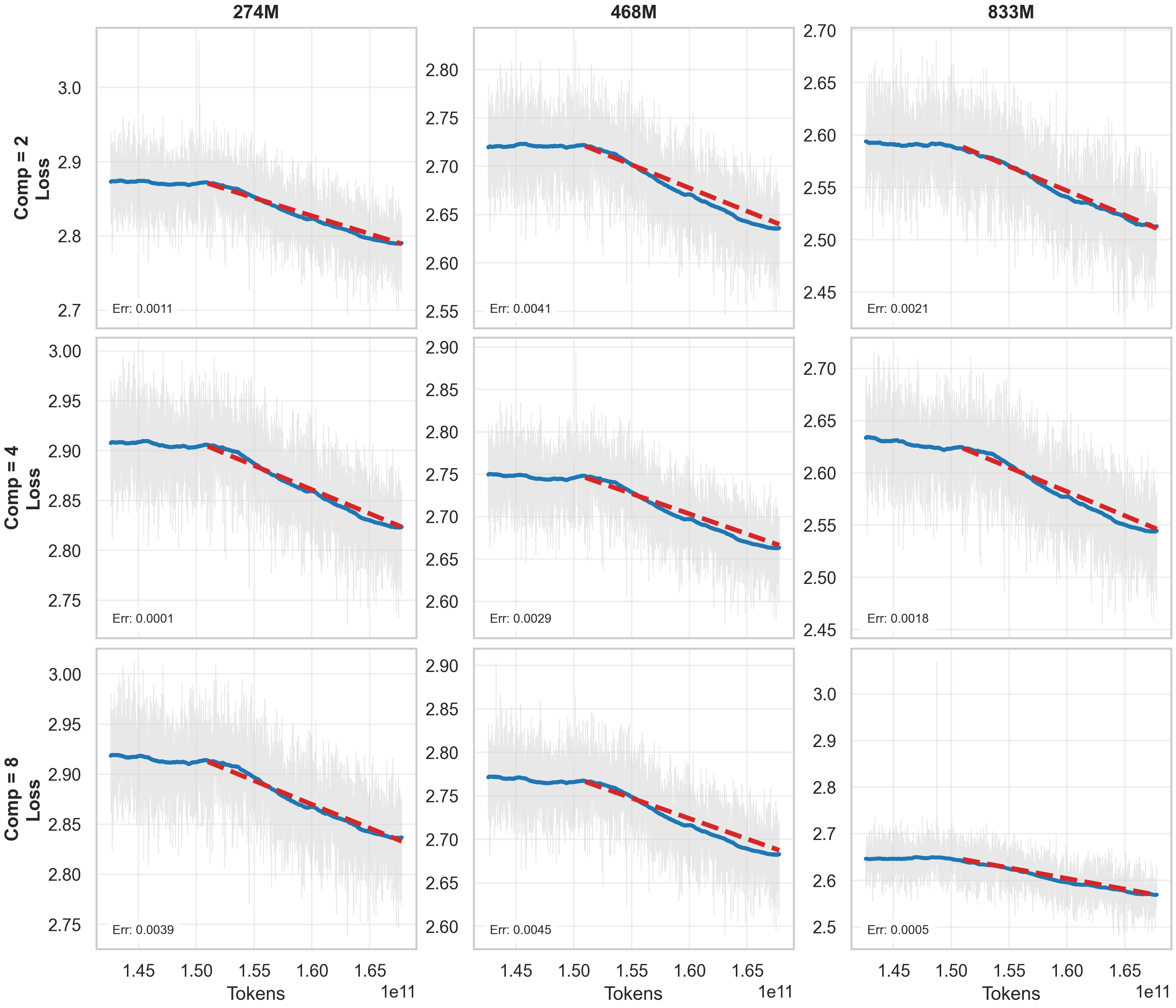}
  \caption{\textbf{Decay-phase fit.} Simplified fit on the final portion of training tokens, validating that our WSD scaling law accurately captures late-stage behaviour with $R^2 = 0.93$.}
  \label{fig:decay_fit}
\end{figure}

As shown in Figure~\ref{fig:full_training_fit} and Figure~\ref{fig:decay_fit}, our methodology—incorporating a tail-focused sampling strategy and weighting late-token regions—ensures the law generalizes reliably across both architectural and data scales.

\subsection{Optimal Configuration Analysis}

\subsubsection{Architectural Efficiency}
\begin{figure*}[t]
  \centering
  \begin{minipage}{0.48\linewidth}
    \centering
    \includegraphics[width=\linewidth]{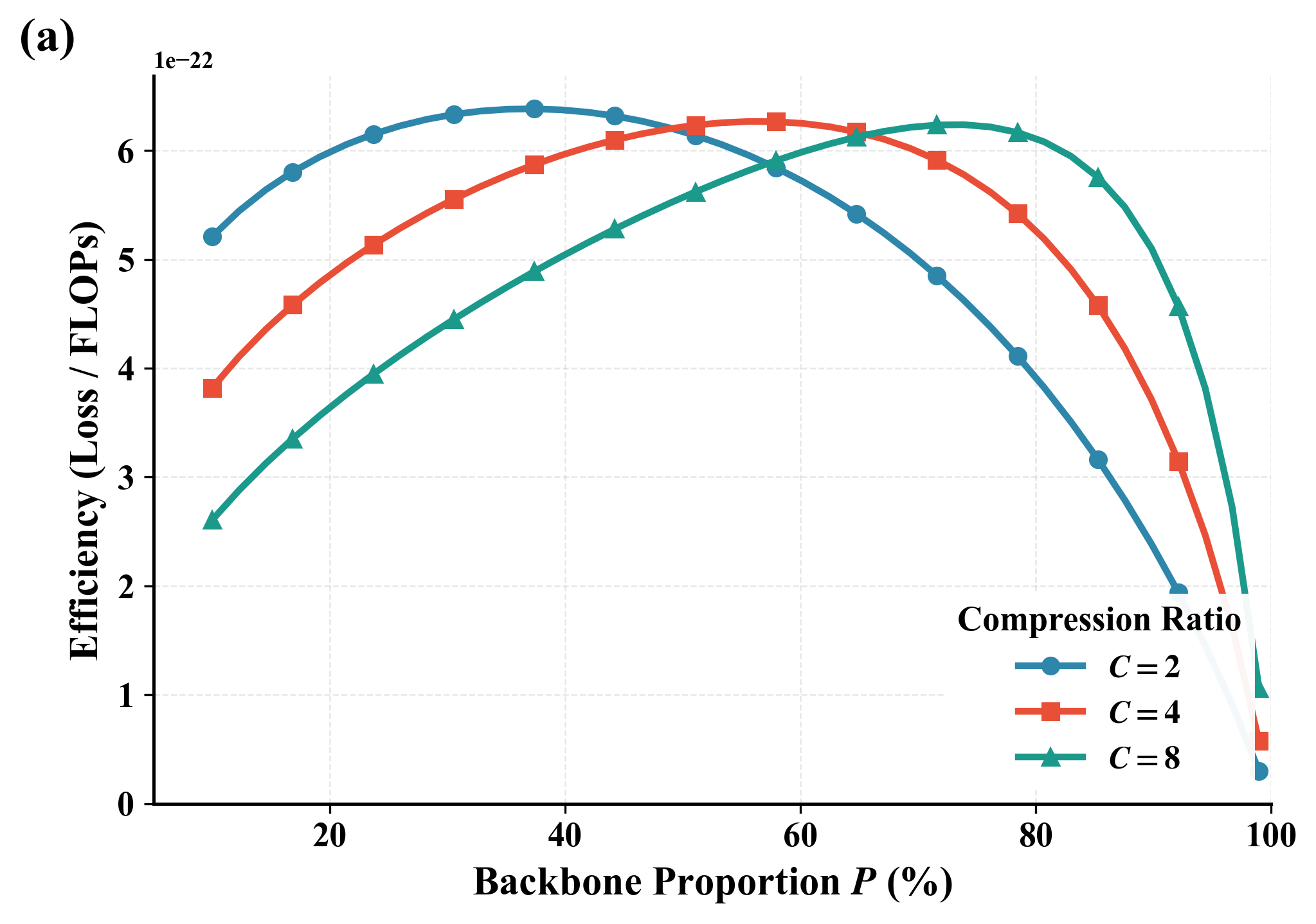}
  \end{minipage}\hfill
  \begin{minipage}{0.48\linewidth}
    \centering
    \includegraphics[width=\linewidth]{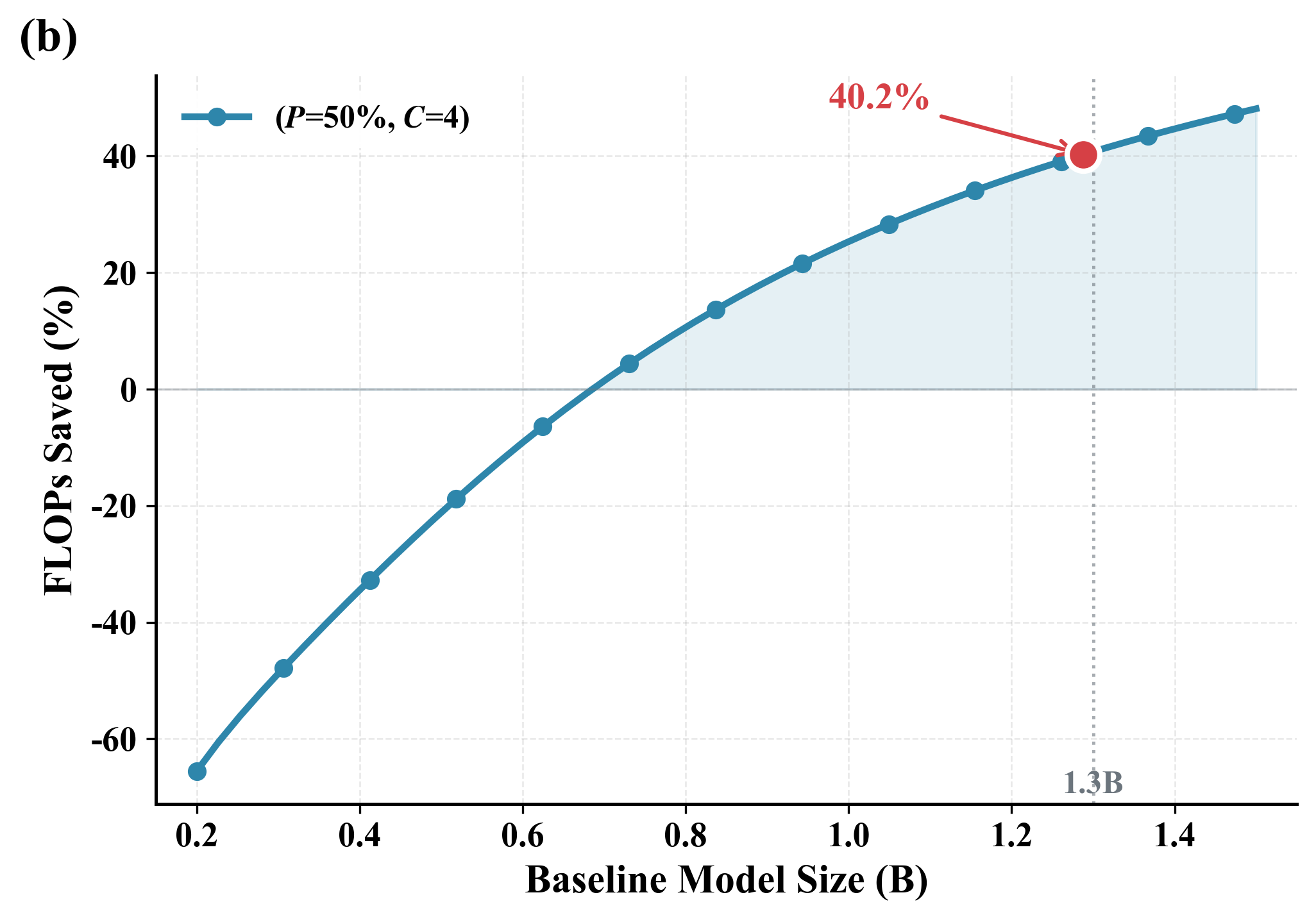}
  \end{minipage}
  \caption{Efficiency analysis of the \ModelName{} architecture. 
(a) Architectural efficiency (Loss/FLOPs) across backbone proportions $P$ 
for different compression ratios $R$. (b) FLOPs savings compared to baseline 
models of varying sizes, with \ModelName{} configured at $P=60\%, R=4$.}
  \label{fig:wsd_scaling_law_fit}
\end{figure*}

Figure~\ref{fig:wsd_scaling_law_fit}(a) illustrates the Loss/FLOPs efficiency across different backbone proportions $P$ and compression ratios $R$.

\paragraph{Selection of Compression Ratio ($R=4$):}
While higher compression rates offer theoretical FLOPs savings, we empirically selected $R=4$ as the primary configuration. This decision is driven by the granularity of concept compression; as discussed in Appendix~\ref{appendix:segmentation}, a compression ratio of 4 aligns better with the intuitive semantic segmentation of tokens into concepts, offering the best balance between training stability and computational efficiency.

\subsection{Scaling Law Validation and Verification}

We developed a unified scaling-law estimator by jointly modeling the full-training loss trajectory and the late-stage decay behavior. To ensure robustness, we adopted a tail-focused sampling strategy that emphasizes curvature near convergence, performing fits on 100B-token trajectories while validating against 1T-token limits. As shown in Figure~\ref{fig:full_training_fit} and Figure~\ref{fig:decay_fit}, our estimator maintains a fitting error below 0.05 across the entire window. 

This high-fidelity fitting allows for a critical verification of our architectural properties: our scaling law yields an effective compute multiplier prediction of approximately \textbf{1.4}. This value aligns closely with the standard baseline factor of \textbf{1.34}. This consistency confirms that our theoretical projections are grounded in established empirical norms and that the architecture scales predictably under the proposed law.


\section{Experiments}

\subsection{Main Results}

We compare \ModelName{} against a parameter-matched baseline that
follows the LLaMA~\cite{touvron2023llama} architecture. Both models are
trained from scratch on our proprietary dataset, using the same global
batch size, learning rate, and sequence length as reported in the LLaMA
paper. Each model is trained on 1T tokens. Results on 12 standard
zero-shot benchmarks are summarized in
Table~\ref{tab:categorized_comparison_step238000}.

Our model follows an \emph{encoder--compressor--decoder} architecture with learned concept circulation, explicitly redistributing computation from uniform token-level processing to adaptive concept-level reasoning.
As a consequence, we do not expect uniform gains across all benchmarks.
Instead, performance differences directly reflect the architectural bias induced by semantic compression and boundary-aware compute allocation.

Overall, \ModelName{} achieves an average accuracy of \textbf{43.92\%}, surpassing the baseline score of 41.23\% by \textbf{+2.69\%}.
However, these gains are highly non-uniform across tasks, revealing a clear separation between reasoning-dominant benchmarks and those that rely on fine-grained token-level alignment.

\paragraph{Reasoning-Dominant Tasks.}
We observe consistent and often substantial improvements on benchmarks that emphasize multi-step reasoning, hypothesis selection, and implicit commonsense inference.
Notable gains are achieved on \textbf{CommonsenseQA (+1.64\%)}, \textbf{HellaSwag (+0.67\%)}, \textbf{OpenBookQA (+3.00\%)}, \textbf{PIQA (+2.42\%)}, and both \textbf{ARC Easy (+2.61\%)} and \textbf{ARC Challenge (+1.77\%)}.
These tasks are characterized by non-uniform information density, where prediction difficulty concentrates around semantic transitions rather than being evenly distributed across tokens.
By compressing locally predictable spans and allocating the majority of model capacity to a high-dimensional concept backbone, \ModelName{} focuses computation on structurally salient regions.
This behavior is consistent with our loss distribution analysis in Section~7.2, which shows systematic loss reduction near concept boundaries.

\paragraph{Granularity-Sensitive Text Understanding.}
In contrast, we observe mild regressions on \textbf{BoolQ (-1.47\%)} and \textbf{RACE (-0.72\%)}.
These benchmarks depend heavily on fine-grained sentence-level entailment, polarity resolution, and subtle lexical cues.
The encoder--compress--decode paradigm inevitably reduces token-level granularity within concept interiors, which can obscure micro-level distinctions required for such tasks.
Importantly, this degradation is localized rather than uniform: while boundary tokens are modeled more accurately, mid-concept positions may trade off fine-grained precision for improved global coherence.
This trade-off manifests as the U-shaped loss profile observed in our mechanistic analysis.

\paragraph{Knowledge and Multilingual Benchmarks.}
For encyclopedic knowledge evaluation, we observe mixed behavior.
While \textbf{C-Eval (+1.71\%)} benefits from adaptive segmentation enabled by the Global Parser, slight regressions appear on \textbf{MMLU (-0.30\%)} and \textbf{CMMLU (-0.24\%)}.
These datasets reward relatively uniform factual recall across tokens, leaving less opportunity for boundary-aware compute reallocation.
This result further supports our central claim: \ModelName{} is structurally optimized for reasoning under non-uniform information density, rather than uniform memorization-heavy retrieval.

\paragraph{Architecture and Parameter Efficiency.}
Although \ModelName{} contains nearly \textbf{2$\times$} the total parameters of the baseline model (2.3B vs.\ 1.3B), this increase is deliberately concentrated in the concept-level backbone ($d_{\text{concept}}=3072$).
Because this backbone operates on a sequence compressed by $4\times$, the effective FLOPs per inference step remain comparable to the smaller baseline.
This validates our core design principle: shifting computation from redundant token-level processing to dense concept-level reasoning enables substantially larger effective capacity without incurring proportional inference cost.





\begin{table}[htbp]
\centering
\caption{\textbf{Performance Comparison: DLCM vs. Baseline.} Zero-shot accuracy (\%) categorized by task type. Improvements are shown in \textcolor{green}{green} and regressions in \textcolor{red}{red}.}
\label{tab:categorized_comparison_step238000}
\renewcommand{\arraystretch}{1.2}
\setlength{\tabcolsep}{10pt}
\begin{tabular}{lccc}
\toprule
\rowcolor{gray!15}
\textbf{Task / Category} & \textbf{DLCM (Ours)} & \textbf{Baseline} & \textbf{Diff.} \\
\midrule
\multicolumn{4}{l}{\textit{Multi-choice General Knowledge / Common Sense}} \\

Commonsense QA  & \textbf{21.38} & 19.74 & \textcolor{green}{+1.64} \\
HellaSwag       & \textbf{46.66} & 45.99 & \textcolor{green}{+0.67} \\
Winogrande      & \textbf{57.22} & 56.20 & \textcolor{green}{+1.02} \\
OpenBookQA      & \textbf{26.80} & 23.80 & \textcolor{green}{+3.00} \\
PIQA            & \textbf{75.52} & 73.10 & \textcolor{green}{+2.42} \\
ARC Challenge   & \textbf{34.81} & 33.04 & \textcolor{green}{+1.77} \\
ARC Easy        & \textbf{69.91} & 67.30 & \textcolor{green}{+2.61} \\
MMLU            & 25.40 & \textbf{25.70} & \textcolor{red}{-0.30} \\
\midrule
\multicolumn{4}{l}{\textit{Multi-choice Text Understanding}} \\
BoolQ           & 62.54 & \textbf{64.01} & \textcolor{red}{-1.47} \\
RACE            & 35.31 & \textbf{36.03} & \textcolor{red}{-0.72} \\
\midrule
\multicolumn{4}{l}{\textit{Culture / Multilingual Knowledge}} \\
C-Eval          & \textbf{26.08} & 24.37 & \textcolor{green}{+1.71} \\
CMMLU           & 25.23 & \textbf{25.47} & \textcolor{red}{-0.24} \\
\midrule
\rowcolor{gray!15}
\textbf{Average} & \textbf{43.92} & 41.23 & \textcolor{green}{\textbf{+2.69}} \\
\bottomrule
\end{tabular}
\end{table}

\begin{table}[ht]
\centering
\caption{\textbf{Architecture Configuration Details.} A unified view of the parameter settings for Baseline (LLaMA-1.3B) and \ModelName{} (2.3B). Values are presented as \textit{Baseline / Ours}.}
\label{tab:architecture_comparison}
\small
\renewcommand{\arraystretch}{1.2} 
\setlength{\tabcolsep}{5pt}       

\begin{tabular}{ll @{\hspace{1.2cm}} ll} 
    \toprule
    \multicolumn{2}{l}{\textbf{General Settings}} & \multicolumn{2}{l}{\textbf{Dimension Settings}} \\
    \cmidrule(r){1-2} \cmidrule(l){3-4} 
    \textit{Metric} & \textit{Value} & \textit{Metric} & \textit{Value} \\
    \midrule
    Model Type      & Trans. / \textbf{DLCM}    & Hidden Size ($d_{\text{token}}$)   & 1,536 \\
    Total Params    & 1.3B / \textbf{2.3B}      & Main Hidden ($d_{\text{concept}}$) & -- / \textbf{3,072} \\
    Vocab Size      & 128,815                   & Interm. Size (Self)                & 4,096 / 6,144 \\
    Max Pos Emb     & 8k /8k         & Interm. Size (Cross)               & -- / 6,144 \\
    Activation      & Swish                     &                                    & \\ 
    \midrule
    
    \multicolumn{2}{l}{\textbf{Layer Configuration}} & \multicolumn{2}{l}{\textbf{Attention Configuration}} \\
    \cmidrule(r){1-2} \cmidrule(l){3-4}
    Total Layers    & 32                        & Attn Heads      & 24 / 24 \\
    Encoder Layers  & -- / \textbf{10}          & Backbone Heads  & -- / \textbf{48} \\
    Backbone Layers & -- / \textbf{16}          & KV Heads        & 24 / 12 \\
    Decoder Layers  & -- / \textbf{6}           & Backbone KV     & -- / \textbf{24} \\
    \bottomrule
\end{tabular}
\end{table}

\subsection{Analysis: Compute Allocation in Concept-Based Models}

\subsubsection{Experimental Setup}
To isolate the impact of concept-based compression on model behavior, we conduct a controlled comparison between our proposed concept model and a standard Transformer baseline. Both models utilize the same backbone architecture (1.3B parameters) and were trained on an identical subset of $100$B tokens from our pretraining corpus. This ensures that any observed differences in loss distribution are attributable solely to the compression mechanism and architectural changes, rather than discrepancies in training data or compute budget.

\subsubsection{Loss Distribution Analysis}
To understand how the model allocates computational resources, we evaluate the loss distribution across relative positions within concepts. We randomly selected 600 samples from the validation set and aligned the token-level losses based on their position within a segmented concept (e.g., the $i$-th token of a concept).

\begin{figure}[h]
    \centering
    \includegraphics[width=0.8\textwidth]{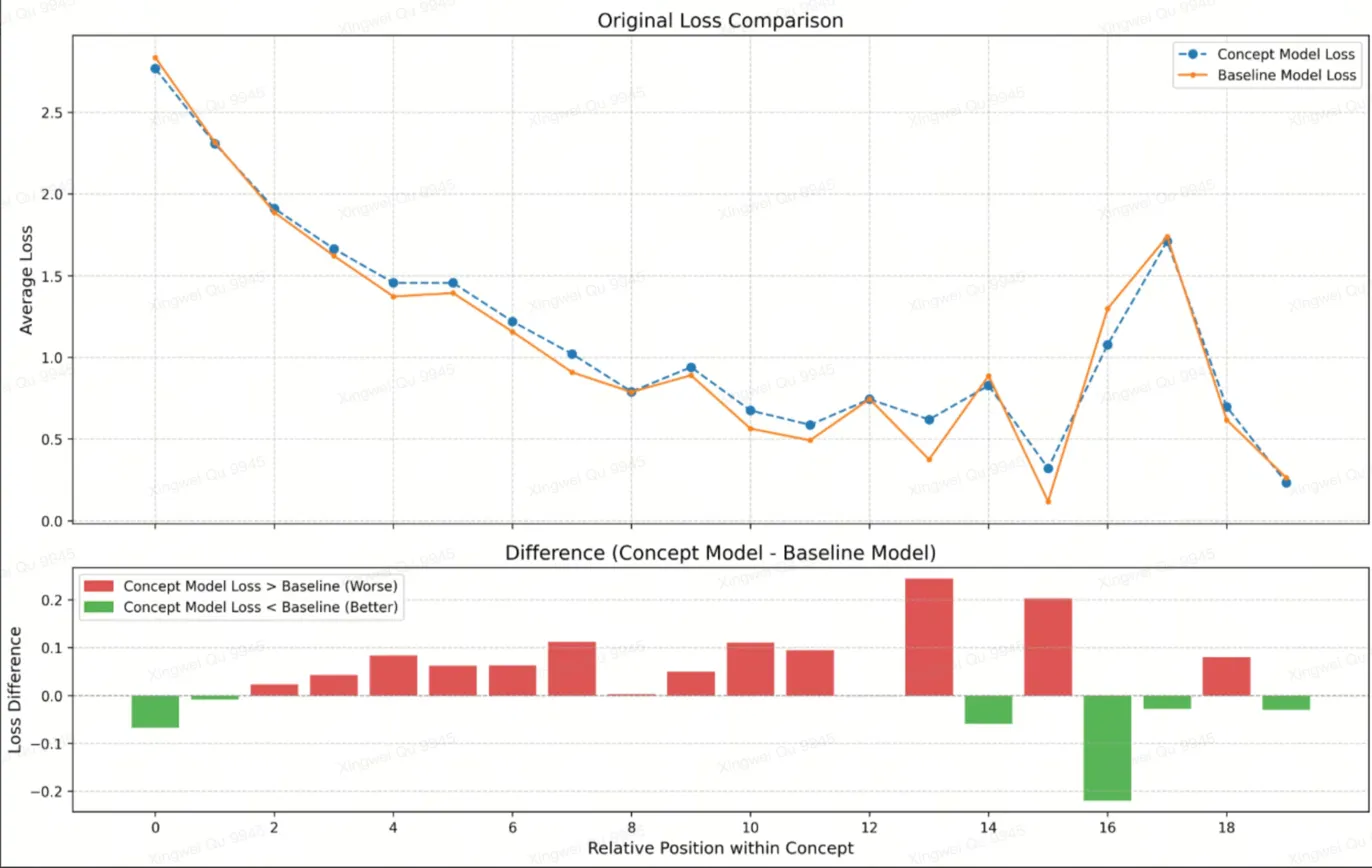}
    \caption{Top: Average loss comparison between concept model (blue) and baseline model (orange) across relative positions within concepts. Bottom: Loss difference (Concept - Baseline), where \textbf{green indicates improvement} (lower loss) and red indicates degradation.}
    \label{fig:loss_comparison}
\end{figure}

Figure~\ref{fig:loss_comparison} illustrates the average loss at the first 20 positions within each concept. The top panel compares raw loss values, while the bottom panel visualizes the differential: $\Delta L = L_{\text{concept}} - L_{\text{baseline}}$. Here, \textbf{green bars} ($\Delta L < 0$) indicate the concept model outperforms the baseline, while \textbf{red bars} ($\Delta L > 0$) indicate degradation.

The results reveal a distinct "U-shaped" improvement pattern that reflects the model's resource reallocation strategy:

\begin{enumerate}
    \item \textbf{Boundary Proficiency (Positions 0--2 \& 16+):} Consistent with the observation that concept models excel at initial and late positions (indicated by green bars), the architecture effectively captures the transition semantics. By explicitly modeling concept boundaries, the model reduces ambiguity at the start and end of semantic units, outperforming the baseline which treats these tokens uniformly.
    
    \item \textbf{Internal Complexity (Mid-positions):} In the middle of a concept (approx. positions 4--15), we observe a shift. While the baseline model often struggles here (higher absolute loss), the concept model's performance is mixed. The presence of red bars in certain mid-concept regions suggests that the compression mechanism forces the model to trade off some fine-grained token-level precision to maintain higher-level semantic coherence.
\end{enumerate}

This reallocation aligns with our hypothesis: the concept model sacrifices uniform token-level predictability (resulting in minor degradation at specific internal positions) to gain superior performance at semantic boundaries and structurally critical tokens. This strategic trade-off allows the model to "spend" its capacity on maintaining global coherence, explaining the downstream improvements despite non-uniform loss reduction.

\section{Ablation Studies}

\subsection{Analysis: End-to-End Discrete Boundary Learning vs. Decoupled Segmentation}

This experiment compares two boundary prediction mechanisms for sequence compression: a learned neural predictor with compression rate regularization (Section~\ref{sec:adaptive_compression}) and a rule-based predictor using cosine similarity. Starting with sequences of length $L=8192$, we track the average compressed length during training.

\begin{figure}
    \centering
    \includegraphics[width=0.8\textwidth]{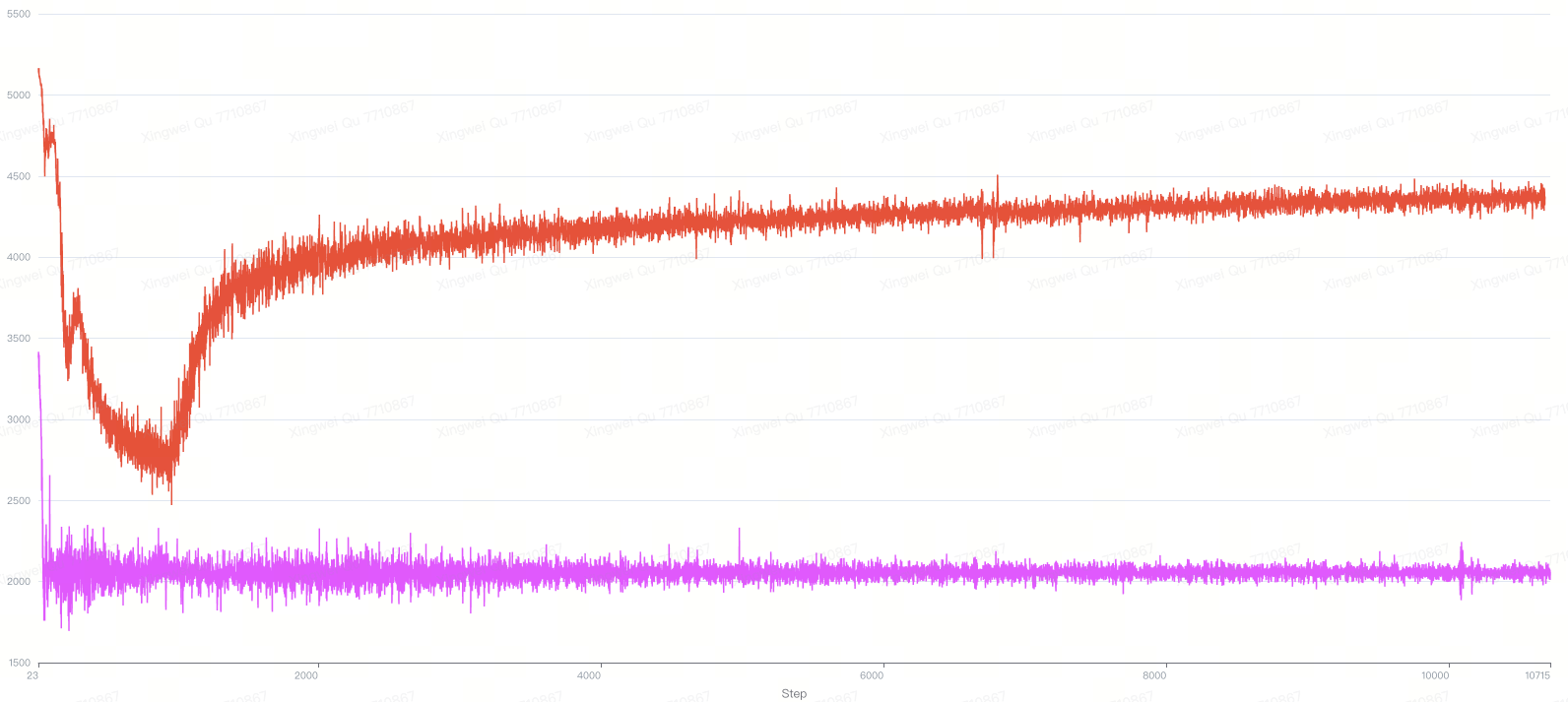}
    \caption{Average compressed sequence length over training steps. \textbf{Red:} Learned Boundary Predictor. \textbf{Purple:} Rule-Based Predictor. The x-axis represents training steps, and the y-axis represents the average number of tokens post-compression.}
    \label{fig:comparison}
\end{figure}

Figure \ref{fig:comparison} reveals starkly different behaviors. The \textbf{learned predictor (red)} exhibits severe instability: after initial compression to $\sim$2000 tokens, the compressed length steadily increases, eventually stabilizing at $\sim$4300 tokens ($1.9\times$ compression). This "creep-up" indicates the model progressively learns to compress less over time. In contrast, the \textbf{rule-based predictor (purple)} demonstrates exceptional stability, rapidly converging to $\sim$2000 tokens ($4\times$ compression) and maintaining this level consistently throughout training.

The learned predictor's instability stems from conflicting optimization objectives. Despite the compression rate regularization term $\mathcal{L}_{\text{aux}}$ designed to maintain the target compression ratio $R$, the primary cross-entropy (CE) loss creates much stronger gradients that penalize information loss and discourage compression:
\begin{equation}
\nabla_{\theta} \mathcal{L}_{\text{total}} = \underbrace{\nabla_{\theta} \mathcal{L}_{\text{CE}}}_{\text{anti-compression}} + \lambda \underbrace{\nabla_{\theta} \mathcal{L}_{\text{aux}}}_{\text{pro-compression}}
\end{equation}
Since $\|\nabla_{\theta} \mathcal{L}_{\text{CE}}\| \gg \lambda \|\nabla_{\theta} \mathcal{L}_{\text{aux}}\|$, the CE loss eventually dominates, forcing the predictor to reduce segmentation despite the regularization term.

The rule-based predictor avoids this conflict through a fixed decision rule: $p_t = \frac{1 - \cos(h_t, h_{t+1})}{2}$, with boundaries inserted when $p_t > \tau$. While the representations $h_t$ are learned, the segmentation rule itself is not optimized by the CE loss. This decoupling prevents the task loss from undermining the compression mechanism, ensuring stable and controllable compression ratios through the threshold parameter $\tau$.

\subsection{Global Regularization via Gradient Accumulation}

To further stabilize the learned boundary predictor, we investigate an alternative regularization strategy: computing the compression ratio loss over accumulated training examples rather than individual sequences (Section~\ref{sec:adaptive_compression}). This \textbf{global regularization} approach computes boundary statistics $F_{\text{global}}$ and $G_{\text{global}}$ across all tokens in $K$ micro-batches.

We train two 2.3B parameter models for 1T tokens with a target compression ratio of $R=2$: one with per-sequence regularization ("Normal") and one with global regularization ("Global Parser"). Table \ref{tab:ablation} presents the downstream performance and the actual realized compression ratios.

\begin{table}[htbp]
\centering
\caption{\textbf{Ablation Study: Global Parser vs. Normal.} Performance comparison on downstream tasks. Both models aim for a target compression ratio of $R=4$. The \textbf{Global Parser} achieves a realized ratio much closer to the target while consistently improving accuracy on most tasks.}
\label{tab:ablation}
\renewcommand{\arraystretch}{1.15} 
\setlength{\tabcolsep}{10pt}       

\begin{tabular}{lccc}
\toprule
\rowcolor{tabgray}
\textbf{Task} & \textbf{Global Parser} & \textbf{Normal} & \textbf{Metric} \\
\midrule
ARC Challenge   & \textbf{0.3038} & 0.2858 & Acc \\
ARC Easy        & \textbf{0.6296} & 0.6242 & Acc \\
Commonsense QA  & \textbf{0.2457} & 0.2228 & Acc \\
HellaSwag       & \textbf{0.3507} & 0.3499 & Acc \\
OpenBookQA      & 0.3220 & \textbf{0.3280} & Acc \\
PIQA            & \textbf{0.6806} & 0.6785 & Acc \\
\midrule
\rowcolor{tabgray}
\textbf{Avg. Improvement} & \textcolor{green}{\textbf{+2.1\%}} & -- & -- \\
\textbf{Realized Ratio}   & \textbf{3.92} & 3.15 & (Target $R{=}4$) \\
\bottomrule
\end{tabular}
\end{table}

The global regularization approach achieves consistently better performance across most tasks (5 out of 6). Crucially, as shown in the bottom row of Table~\ref{tab:ablation}, the Global Parser maintains a realized compression ratio ($\sim$3.9) much closer to the target ($4.0$) compared to the Normal formulation, which tends to degrade towards lower compression.

The key insight is that enforcing a fixed compression ratio per sequence is overly restrictive. Real-world data exhibits varying information density. By relaxing the constraint to operate at the batch level, the global regularization allows the model to learn adaptive behavior—compressing repetitive code more aggressively while preserving dense technical text—effectively allocating the compression "budget" where it matters most.

\subsection{Content-Adaptive Compression Benefits}

To verify the adaptive behavior enabled by global regularization, we analyzed the segmentation granularity across different domains. Table~\ref{tab:compression_stats} presents empirical measurements of the average tokens per concept.

\begin{table}[h]
\centering
\begin{tabular}{lccc}
\hline
\textbf{Content Type} & \textbf{Target 8×} & \textbf{Target 4×} & \textbf{Target 2×} \\
\hline
Casual English    & 7.47 & 3.53 & 1.76 \\
Casual Chinese    & 8.38 & 4.36 & 1.76 \\
Technical English & 10.58 & 3.85 & 1.92 \\
Technical Chinese & 6.09 & 3.27 & 1.76 \\
Code              & 6.14 & 3.66 & 1.98 \\
Math/Science      & 7.42 & 4.41 & 1.91 \\
\hline
\end{tabular}
\caption{Average tokens per concept across content types and compression ratios. Values represent the actual granularity achieved for each target compression setting.}
\label{tab:compression_stats}
\end{table}

The data reveals significant variation in compression density across content types. For instance, at the 8× target, Technical English retains significantly more tokens per concept (10.58) compared to Technical Chinese (6.09) or Code (6.14). 

While the precise ranking of "optimal" length varies across compression targets (as noted in the fluctuation between content types), the \textbf{existence of this variation} is the critical finding. It confirms that the global regularization mechanism successfully decouples the compression objective from rigid per-sequence constraints. The model is not forcing a uniform segment length; instead, it adapts the granularity based on the inherent semantic density of the content. Code and structured text tend to be compressed into shorter, syntactic units, whereas dense prose is preserved in longer semantic chunks. This adaptivity—regardless of the specific order—allows the model to maximize information retention within the global compression budget.
\section{Conclusion}

We presented \textbf{Dynamic Large Concept Models (DLCM)}, a hierarchical language modeling framework that challenges the token-uniform computation paradigm underlying modern LLMs. By learning semantic boundaries from latent representations and shifting computation from tokens to variable-length concepts, DLCM enables reasoning to occur in a compact, semantically aligned space rather than repeatedly at the token level.

Beyond the architectural design, we showed that hierarchical compression necessitates new theoretical and optimization tools. We introduced a compression-aware scaling law that clarifies how compute should be allocated between token processing and concept-level reasoning under fixed FLOPs, and developed a decoupled $\mu$P parametrization that enables stable training and zero-shot hyperparameter transfer in heterogeneous architectures. Empirically, DLCM achieves consistent gains on reasoning-intensive benchmarks while reducing redundant computation, demonstrating a favorable accuracy--efficiency trade-off.

More broadly, our results suggest that scaling language models is not solely a matter of increasing parameters or data, but also of reconsidering \emph{where} computation is performed. We believe concept-level latent reasoning offers a promising direction for building more efficient and more reasoning-capable language models, and opens avenues for future work on adaptive abstraction, planning, and multi-level reasoning in large-scale neural systems.

\newpage
\section*{Contributions}

\textbf{Leading Authors}

Xingwei Qu, Shaowen Wang, Zihao Huang, Ge Zhang

\textbf{Leading Author Contributions}

Xingwei Qu: Conducts fundamental ablation studies and is responsible for the majority of the engineering implementation.

Shaowen Wang: Implements and advances the MuP (Maximal Update Parametrization) for hyperparameter tuning.

Zihao Huang: Implements noise based boundary prediction tricks.

Ge Zhang: Proposes the original idea and develops the demo prototype. Identifies and provides the key technique for the Global Parser.

\textbf{Core Contributors}

Kai Hua: Designs and constructs the training data entirely from open-source data.

Fan Yin: Contributes to the SGLang implementation and optimization.

Rui-Jie Zhu: Resolves bugs related to \ModelName{} and proposes the strategy of replacing the encoder with \ModelName{}.

Jundong Zhou \& Qiyang Min: Contributes to the project development and implementation.

\textbf{Other Contributors}

Zihao Wang, Yizhi Li, Tianyu Zhang, He Xing, Zheng Zhang, Yuxuan Song, Tianyu Zheng, Zhiyuan Zeng

\textbf{Corresponding Authors}

Chenghua Lin, Ge Zhang, Wenhao Huang

\clearpage

\bibliographystyle{plainnat}
\bibliography{main}

\clearpage

\beginappendix

\section{Appendix: Segmentation Examples at Different Compression Ratios}
\label{appendix:segmentation}

We provide representative examples of how boundary prediction behaves under different compression ratios across three content types: casual English text, Python code, and mathematical exposition.

\subsection{Casual English Text}

\textbf{Original (90 tokens):}
\begin{quote}
\small
So I've been trying to perfect my morning coffee routine lately. It's funny how something so simple can have so many variables. I started with a basic drip machine, which was fine, but a bit boring. Then I went down the French press rabbit hole – way more flavor, but the cleanup is a real hassle, you know?
\end{quote}

\textbf{Compression 8× (11 segments):}
\begin{quote}
\small
So I | 've been trying to perfect my morning coffee routine lately. | It's funny how something | so simple can have so many variables. | I started with | a basic drip machine, which | was fine, but a | bit boring. | Then I went down the French press rabbit hole – | way more flavor, but the cleanup is a | real hassle, you know?
\end{quote}

\textbf{Compression 4× (35 segments):}
\begin{quote}
\small
So I | 've been trying | to perfect | my morning coffee routine lately. | It's | funny how something | so simple can | have so many variables. | I started | with | a basic drip machine, which was fine, but | a bit boring.
\end{quote}

\textbf{Compression 2× (56 segments):}
\begin{quote}
\small
So I | 've | been trying to | perfect | my | morning coffee routine | lately. | It | 's funny how | something so | simple | can | have so many | variables.
\end{quote}

\subsection{Python Code}

\textbf{Original (87 tokens):}
\begin{verbatim}
import torch
from torch.utils.data import Dataset, DataLoader
class SimpleTextDataset(Dataset):
    """A simple dataset for loading text data."""
    def __init__(self, texts, tokenizer, max_length=128):
        self.texts = texts
        self.tokenizer = tokenizer
\end{verbatim}

\textbf{Compression 8× (12 segments):}
\begin{verbatim}
import torch | from torch.utils.data import Dataset, DataLoader | 
class SimpleTextDataset(Dataset): | 
"""A simple dataset for loading text data.""" | 
def __init__(self, texts, tokenizer, max_length=128): | 
self.texts = | texts | self.tokenizer = | tokenizer
\end{verbatim}

\textbf{Compression 4× (29 segments):}
\begin{verbatim}
import torch | from torch.utils.data | import Dataset | , DataLoader | 
class Simple | TextDataset(Dataset): | """ | A | simple dataset for | 
def | __init__(self, texts, tokenizer, max_length= | 128): | 
self.texts = | texts
\end{verbatim}

\textbf{Compression 2× (58 segments):}
\begin{verbatim}
import | torch | from torch.utils.data import | Dataset, | DataLoader | 
class | Simple | TextDataset(Dataset): | """ | A simple | dataset for | 
def | __init__(self, | texts, | tokenizer | , | max_length= | 128): | 
self | .texts | = | texts
\end{verbatim}

\subsection{Mathematical Text}

\textbf{Original (95 tokens):}
\begin{quote}
\small
Euler's formula is a mathematical formula in complex analysis that establishes the fundamental relationship between the trigonometric functions and the complex exponential function. The formula states that for any real number x, $e^{ix} = \cos(x) + i\sin(x)$, where e is the base of the natural logarithm, i is the imaginary unit.
\end{quote}

\textbf{Compression 8× (13 segments):}
\begin{quote}
\small
Euler's formula is a | mathematical formula in complex analysis that establishes | the fundamental relationship between the trigonometric functions and the complex exponential function. | The formula states that | for any real number x, $e^{ix}$ = | $\cos(x) + i\sin(x)$, | where e | is the | base of the natural logarithm, i is the | imaginary unit
\end{quote}

\textbf{Compression 4× (32 segments):}
\begin{quote}
\small
Euler's | formula is | a mathematical formula in | complex analysis that establishes | the fundamental relationship between | the trigonometric functions and the | complex exponential function. | The formula states | that for any | real number x | , $e^{ix}$ = | $\cos(x) + i\sin(x)$, | where e | is | the base of the natural logarithm
\end{quote}

\textbf{Compression 2× (61 segments):}
\begin{quote}
\small
Euler | 's | formula is | a | mathematical formula | in | complex analysis that | establishes | the fundamental relationship between | the trigonometric functions | and | the complex exponential | function. | The | formula states | that | for | any | real number x | , | $e^{ix}$ = | $\cos(x)$ | + | $i\sin(x)$,
\end{quote}

\begin{table}[h]
\centering
\caption{Performance comparison (Batch=1, Heads=32, Interval=6)}
\label{tab:perf_comparison}
\begin{tabular}{@{}ccccc@{}}
\toprule
Seq Length & Hidden Size & Flex (ms) & Flash Varlen (ms) & \textbf{Speedup} \\
\midrule
2048  & 1024 & 32.35 & 22.48 & \textbf{1.44$\times$} \\
2048  & 2048 & 33.31 & 22.58 & \textbf{1.48$\times$} \\
2048  & 4096 & 32.42 & 22.56 & \textbf{1.44$\times$} \\
\midrule
4096  & 1024 & 59.75 & 45.15 & \textbf{1.32$\times$} \\
4096  & 2048 & 60.72 & 45.17 & \textbf{1.34$\times$} \\
4096  & 4096 & 65.88 & 45.48 & \textbf{1.45$\times$} \\
\midrule
8192  & 1024 & 116.35 & 91.42 & \textbf{1.27$\times$} \\
8192  & 2048 & 114.65 & 90.66 & \textbf{1.26$\times$} \\
8192  & 4096 & 142.75 & 96.79 & \textbf{1.47$\times$} \\
\midrule
16384 & 1024 & 314.35 & 186.21 & \textbf{1.69$\times$} \\
16384 & 2048 & 323.53 & 186.83 & \textbf{1.73$\times$} \\
16384 & 4096 & 315.69 & 190.38 & \textbf{1.66$\times$} \\
\bottomrule
\end{tabular}
\end{table}

\end{document}